\documentclass{article}
\usepackage[letterpaper,margin=1in]{geometry}
\usepackage[T1]{fontenc}
\usepackage[utf8]{inputenc}
\usepackage{microtype}
\usepackage{amsmath}
\usepackage{amssymb}
\usepackage{graphicx}
\usepackage{booktabs}
\usepackage{multirow}
\usepackage{float}
\usepackage[font=small,labelfont=bf]{caption}
\usepackage{placeins}
\usepackage{authblk}
\usepackage{pifont}
\newcommand{\cmark}{\ding{51}}
\newcommand{\xmark}{\ding{55}}
\usepackage{xcolor}
\usepackage[numbers,sort&compress]{natbib}
\usepackage[colorlinks=true,allcolors=blue!55!black]{hyperref}
\hypersetup{
  pdftitle={Predictive Uncertainty for Neural CAE Surrogates},
  pdfauthor={NVIDIA},
  pdfsubject={Uncertainty quantification for geometry-conditioned CAE surrogates},
  pdfkeywords={uncertainty quantification, CAE, Gaussian process, Concrete MC dropout, deep ensemble}
}

\title{Predictive Uncertainty for Neural CAE Surrogates}

\author{Kaustubh Tangsali}
\author{Mohammad Amin Nabian}
\author{Kelvin Lee}
\author{Carmelo Gonzales}
\author{Sanjay Choudhry}

\affil{NVIDIA}

\date{August 2026}

\begin{document}

\maketitle

\begin{abstract}
Neural surrogates can substantially accelerate computer-aided engineering (CAE) workflows, but their use in design requires uncertainty estimates that remain meaningful across varying geometries, spatial prediction fields, and engineering quantities of interest. We investigate how established uncertainty quantification (UQ) approaches behave when adapted to geometry-conditioned neural surrogates. We compare one closed-form and two sampling-based approaches—a Gaussian process (GP)-based method, concrete Monte Carlo (MC) dropout, and deep ensembles—and evaluate them on three large, industry-relevant CAE datasets for external aerodynamics and crash dynamics.

We examine whether predicted uncertainties have credible magnitudes, identify locations with larger prediction errors, respond to unfamiliar inputs, and remain informative for derived engineering quantities. On the DrivAerStar dataset, where all three methods are compared, each generally assigns higher uncertainty to locations with larger prediction errors, and validation-based rescaling brings interval coverage close to nominal on a disjoint in-distribution test set. Results on AirFRANS and automotive crash also show useful error ranking and interval estimates, but the relative performance of the methods changes with the dataset and evaluation criterion. UQ methods and evaluation metrics should therefore be selected based on the intended downstream CAE decision.
\end{abstract}

\section{Introduction}

\begin{figure}[H]
\centering
\makebox[\textwidth][c]{\includegraphics[width=\textwidth]{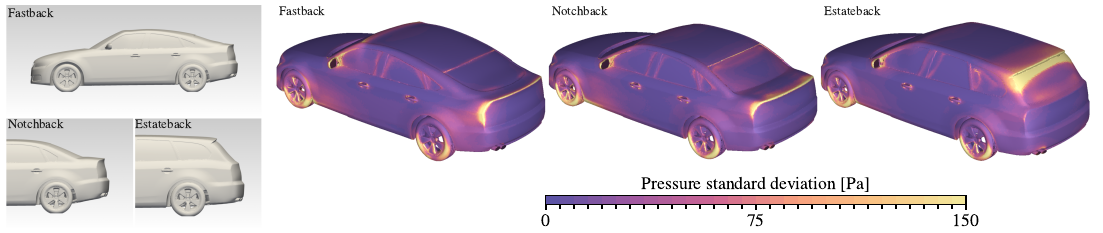}}
\caption{Geometry-conditioned uncertainty across DrivAerStar \citep{qiu2025drivaerstar} body styles. The GP-based surrogate is trained on Fastback geometries and evaluated on held-out Notchback and Estateback classes. Total predictive uncertainty becomes increasingly concentrated around the changing rear-body geometry, with the strongest response on Estateback.}
\label{fig:drivaer-gp-cross-class}
\end{figure}

Neural surrogates reduce the need for repeated high-fidelity simulations in design studies, optimization loops, and interactive engineering analysis. When exhaustive solver verification is impractical, engineers must still decide whether a new geometry warrants a solver run, where its predicted field may be unreliable, and how much confidence to place in drag, intrusion, or another derived quantity. These needs define the three evaluation scopes used here: case level, local-field level, and engineering-quantity level. Figure~\ref{fig:drivaer-gp-cross-class} illustrates the first of these questions for DrivAerStar \citep{qiu2025drivaerstar}, showing the GP uncertainty field when a surrogate trained on Fastbacks is applied to new body styles.

The studies considered here span automotive aerodynamics, high-fidelity airfoil-flow simulation, and transient crash dynamics: large-scale industry-relevant problems involving unstructured meshes, coupled output fields, and variation in geometry, boundary conditions, or structural parameters. This scale and physical diversity make the comparison a test of UQ in representative CAE workloads rather than on a small synthetic regression problem. Although all three evaluation scopes concern uncertainty, their requirements differ: an estimate may be well calibrated yet rank case errors poorly or track local errors without responding to an out-of-distribution (OOD) geometry, while pointwise marginal variances may miss dependencies needed for an integrated quantity. A single calibration score therefore cannot assess all three scopes.

We examine the UQ metrics in a controlled comparison of one closed-form and two sampling-based approaches. The GP-based configuration returns posterior epistemic variance together with a learned residual variance. Concrete MC dropout and the deep ensemble return field samples whose disagreement estimates epistemic variation in the evaluated implementations. For epistemic tasks, GP posterior variance and sampling disagreement are directly comparable estimates of model uncertainty; for predictive intervals, GP total variance includes an additional residual component. In this study, the DrivAerStar \citep{qiu2025drivaerstar} dataset supplies the complete three-method comparison, the AirFRANS \citep{bonnet2022airfrans} dataset changes the shift from vehicle body style to angle-of-attack extrapolation, and the automotive crash dataset \citep{nabian2026automotivecrash} changes both the physics and the relevant engineering output.

We characterize what each evaluated UQ method supports, identify what drives its observed strengths, and test whether the resulting conclusions persist across datasets and evaluation scopes.

\paragraph{Contributions}
\begin{enumerate}\itemsep2pt
\item{a CAE-oriented evaluation protocol spanning case, local-field, and engineering-quantity scopes;}
\item{empirical evidence that calibration, error discrimination, response to distribution shift, and usefulness for engineering quantities are distinct UQ properties that can produce different method rankings.}
\item{a controlled empirical comparison across three industrial CAE datasets spanning automotive aerodynamics, airfoil-flow extrapolation, and transient crash dynamics;}
\item{PhysicsNeMo implementations of GP-based, concrete MC dropout, and ensemble uncertainty behind a common prediction interface.}

\end{enumerate}

\section{Related Work}

Deep ensembles and MC dropout are widely used sampling-based approximations to predictive uncertainty in neural networks \citep{lakshminarayanan2017ensembles,gal2016dropout}. Concrete dropout extends MC dropout by approximating its binary dropout masks with a differentiable continuous distribution, allowing the dropout probabilities to be learned jointly with the model parameters \citep{gal2017concrete}. Gaussian processes take a different route: a kernel and posterior distribution define predictive variance directly \citep{rasmussen2006gpml}, while inducing-point approximations and deep kernel learning make this practical on neural/learned representations \citep{titsias2009variational,hensman2015svgp,wilson2016dkl}. Distance-aware alternatives such as spectral-normalized neural Gaussian processes pursue a similar goal, but depend on preserving useful distances in the learned feature space \citep{liu2020sngp,vanamersfoort2021due}. These methods are usually studied separately, leaving engineers without a controlled comparison of what their uncertainty estimates mean on the same large simulation problems.

Predictive uncertainty can be evaluated in several complementary ways. Calibration methods rescale predicted distributions using held-out data \citep{guo2017calibration,kuleshov2018calibrated} so that predicted uncertainty agrees more closely with observed errors or empirical coverage. Proper scoring rules such as negative log predictive density (NLPD) assess the predictive distribution as a whole \citep{gneiting2007scoring}. Selective prediction evaluates whether withholding low-confidence predictions reduces error \citep{geifman2017selective}, while sparsification evaluates whether high uncertainty coincides with large prediction errors \citep{ilg2018uncertainty}. Evaluating uncertainty becomes more difficult under distribution shift because calibration fitted to in-distribution data may not transfer \citep{ovadia2019trust}. Conformal prediction constructs intervals that attain a chosen coverage rate across new examples without assuming a particular error distribution, provided the calibration and test examples can be treated as draws from the same distribution \citep{angelopoulos2023conformal}. Physics-informed variants use residuals of the governing equations as calibration scores for neural PDE surrogates \citep{gopakumar2025calibrated}. Coverage alone, however, does not show whether uncertainty locates field errors, detects unfamiliar cases, or supports engineering decisions; this is why our evaluation treats those properties separately.

Predictive UQ has been studied in several scientific-computing and CFD settings. Bayesian convolutional encoder--decoder surrogates have been used to predict pressure and velocity fields in porous-media flow \citep{zhu2018bayesian}, and uncertainty-aware convolutional surrogates have been applied to PDE solutions on varying domains \citep{winovich2019convpde}. Bayesian neural networks have also been used within Reynolds-Averaged Navier-Stokes (RANS) workflows to quantify turbulence-model uncertainty and propagate it to flow quantities under changes in geometry and Reynolds number \citep{geneva2019model}. \citet{psaros2023uncertainty} provide a broader review and comparison of UQ methods for PINNs, neural operators, and stochastic PDE problems.

More recent CAE surrogates use neural operators and transformer-based architectures to predict fields on complex geometries \citep{li2021fno,lu2021deeponet,wu2024transolver,luo2025transolverpp,ranade2025domino,adams2025geotransolver}. In particular, DoMINO and GeoTransolver address the irregular geometries and large meshes encountered in industrial simulation. Recent work has also developed UQ methods specifically for neural operators. For example, \citet{magnani2025luno} propagate a linearized weight-space approximation through a trained neural operator to obtain function-valued Gaussian predictions that retain dependencies across output locations. Retaining these dependencies is important when field uncertainty must be propagated through integrals or other operations used to compute engineering quantities \citep{smith2013uq}.

Our focus is complementary: we compare GP-based, concrete MC dropout, and ensemble uncertainty on large, irregular-mesh CAE problems and evaluate their estimates at the local-field, case, and engineering-quantity scopes.

\section{Predictive Uncertainty in CAE Workflows}
\subsection{Uncertainty scopes}
\label{sec:uncertainty-scopes}

\begin{figure}[H]
\centering
\includegraphics[width=0.8\textwidth]{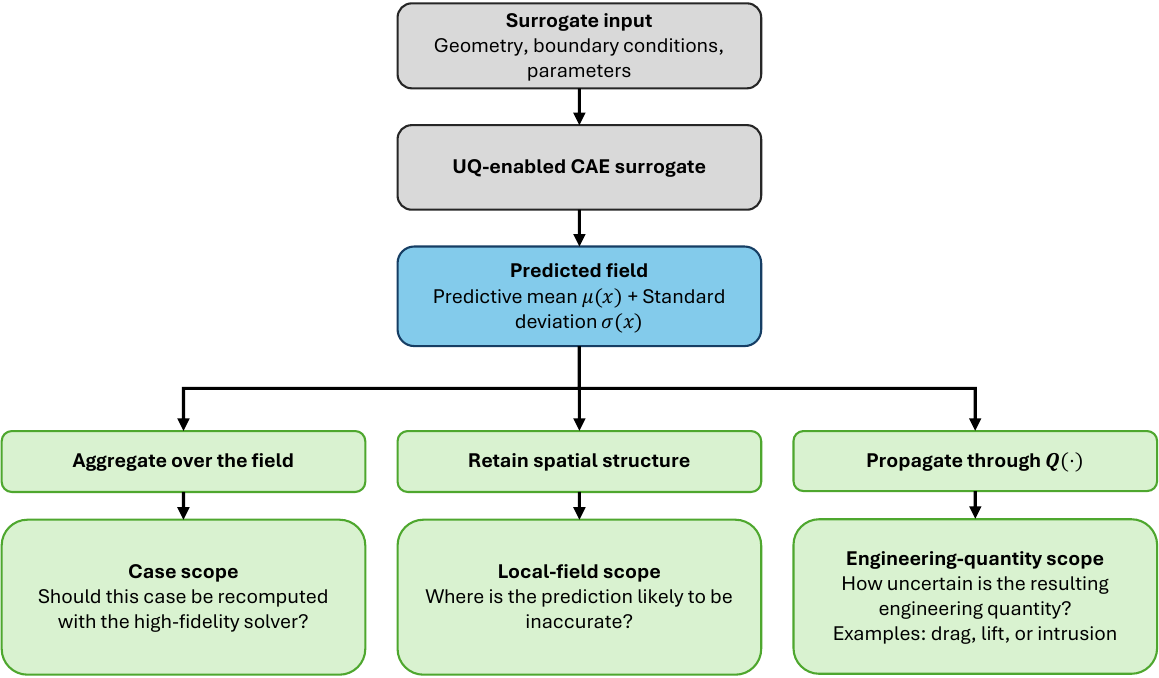}
\caption{Three ways uncertainty is used in the evaluated CAE workflow. Uncertainty in the predicted field can be aggregated to screen a complete input case, retained spatially to identify unreliable regions, or propagated through the calculation of an engineering quantity. These are parallel evaluation scopes rather than a hierarchy.}
\label{fig:uq-scopes}
\end{figure}

We organize the evaluation around three scopes: the case, the local field, and the derived engineering quantity. These scopes describe where uncertainty is used in the engineering workflow, rather than what causes the uncertainty or how it is estimated. Figure~\ref{fig:uq-scopes} summarizes how uncertainty is used at each scope.

At the \emph{case scope}, the question is whether a particular input should be accepted or sent for review with the high-fidelity solver. For a geometry-conditioned surrogate, a case may be difficult because its geometry, boundary conditions, or material configuration differ from those represented in the training data. A case-level score can be formed by aggregating uncertainty over the predicted field and used to prioritize inputs for review or recomputation \citep{geifman2017selective}. An unfamiliar input does not necessarily produce high predicted uncertainty, and a familiar input may still produce a large prediction error. Case-level scores should therefore be tested on both in-distribution and deliberately shifted cases before being used as guardrails \citep{ovadia2019trust}.

At the \emph{local-field scope}, uncertainty is associated with particular spatial locations, time steps, and physical channels. CAE surrogates commonly predict fields such as pressure, velocity, displacement, and stress, so it is useful to know not only whether a complete case is difficult but also where its prediction may be inaccurate. An association between local uncertainty and local prediction error can support error-based triage and help identify regions where additional training data may be valuable \citep{ilg2018uncertainty}. This association is not guaranteed and must be evaluated rather than assumed. For active learning, the epistemic component is generally the relevant quantity because it represents uncertainty that additional training data may reduce \citep{gal2017deepactive}.

At the \emph{engineering-quantity scope}, uncertainty is attached to the quantity on which an engineering decision is ultimately based. Examples include drag and lift coefficients, peak stress, integrated force, and intrusion at a specified location. These quantities are functions of one or more predicted fields, and their uncertainty cannot generally be obtained by independently combining pointwise standard deviations. Spatial correlations and dependencies between physical channels affect how field uncertainty propagates through an integral, extremum, or other engineering calculation \citep{smith2013uq}. Credible uncertainty at the field level therefore does not automatically imply credible uncertainty in the final engineering quantity.

Each scope requires different evaluation criteria. Predictive intervals require credible uncertainty magnitudes, error triage depends on error ranking, guardrails require a reliable response to relevant shifts, and engineering decisions require uncertainty in the final quantity. Section~\ref{sec:metrics} defines the corresponding metrics.

\subsection{Variance Decomposition}
\label{sec:variance-decomposition}

Predictive uncertainty is commonly described in terms of aleatoric and epistemic components \citep{kendall2017uncertainties}. For the GP configuration evaluated here, applying the law of total variance \citep{casella2002statistical} gives

\begin{equation}
\sigma_{\mathrm{pred}}^2(x) = \sigma_{\mathrm{epi}}^2(x) + \sigma_{\mathrm{res}}^2(x),
\end{equation}

where $\sigma_{\mathrm{epi}}^2(x)$ is the posterior variance of the latent predictive function and $\sigma_{\mathrm{res}}^2(x)$ is the likelihood variance conditional on that function \citep{rasmussen2006gpml}.
 
Epistemic uncertainty reflects limited knowledge of the predictive function and may decrease when relevant training data are added. In particular, limited training coverage can leave the predictive response weakly constrained in parts of the input space.

When repeated observations at fixed inputs exhibit genuine variability, $\sigma_{\mathrm{res}}^2(x)$ may have an aleatoric interpretation. The datasets considered here instead contain deterministic simulations, for which a fixed input and solver configuration produce the same target. The predictive mean can nevertheless leave residual errors because the surrogate does not reproduce every feature of the simulated field. The likelihood-fitted residual variance represents the scale of these remaining errors rather than physical randomness, so we interpret it as learned model-discrepancy variance \citep{kennedy2001bayesian}. For the GP-based method, we report this term separately from epistemic variance and include it in the total predictive variance.

\section{Uncertainty Quantification Methods}
We consider one closed-form and two sampling-based approaches. At a high level, the closed-form, GP-based method returns a predictive distribution in one forward pass while sampling methods--concrete MC dropout and deep ensembles--construct empirical distributions from stochastic passes or independently trained models.

\subsection{Gaussian Process-Based Method}
The Gaussian process-based method replaces the deterministic output projection of a neural surrogate with independent per-output variational GPs \citep{titsias2009variational,hensman2015svgp}. The backbone first maps each input point to a learned feature vector, and a deep-kernel-learning network then transforms these features into the representation used by the GP kernel \citep{wilson2016dkl}. Each output channel has its own variational GP, with inducing points used to approximate the posterior.

The GP posterior provides a predictive mean and an epistemic variance. The kernel compares each transformed feature against learned inducing locations, giving the epistemic term an explicit, smooth dependence on distance in the learned feature space. This does not guarantee that epistemic uncertainty will increase under every form of distribution shift. If the backbone or deep-kernel transformation maps unfamiliar inputs into regions already occupied by training features, the GP receives no distance signal from which to increase its posterior variance. Inducing-point coverage, feature normalization, kernel length scales, and preservation of informative feature distances are therefore important parts of the method.

We use a Mat\'{e}rn-$5/2$ kernel with automatic relevance determination (ARD). This kernel provides a finite-smoothness prior and is less restrictive than the infinitely differentiable squared-exponential kernel \citep{rasmussen2006gpml}.
For transformed features $\mathbf{h}$ and $\mathbf{h}'$,
\begin{equation}
k(\mathbf{h},\mathbf{h}')=\sigma_f^2
\left(1+\sqrt{5}r+\frac{5}{3}r^2\right)\exp(-\sqrt{5}r),
\qquad
r^2=\sum_{d=1}^{D}\frac{(h_d-h'_d)^2}{\ell_d^2}.
\end{equation}

Here, $\mathbf{h},\mathbf{h}'\in\mathbb{R}^{D}$ are transformed feature vectors, $d=1,\ldots,D$ indexes their dimensions, $\ell_d$ is the corresponding ARD length scale, and $\sigma_f^2$ is the kernel signal variance. The learned length scales $\ell_d$ allow different sensitivity along each feature dimension. Each physical output channel has an independent kernel, so the GP does not directly model covariance between channels.

For channel $c$, inducing variables $\mathbf{u}_c=f_c(\mathbf{Z}_c)$ have variational posterior $q(\mathbf{u}_c)=\mathcal{N}(\mathbf{m}_c,\mathbf{S}_c)$. The sparse predictive distribution follows from integrating $p(f_c\mid\mathbf{u}_c)$ against $q(\mathbf{u}_c)$. Training minimizes the negative evidence lower bound,
\begin{equation}
\mathcal{L}_{\mathrm{GP}}
=-\sum_{i=1}^{N}\mathbb{E}_{q(f_i)}
\log p(y_i\mid f_i,\sigma_{\mathrm{res},i}^2)
+\beta\,\mathrm{KL}\!\left[q(\mathbf{u})\,\|\,p(\mathbf{u})\right],
\end{equation}
alongside a mean-squared-error anchor used during optimization. The deep-kernel transform is learned jointly with the backbone and GP parameters.

In addition to the posterior epistemic variance, the head contains a learned residual variance. With a homoscedastic likelihood, one residual scale is learned for each output channel. With the heteroscedastic configuration used in this study, a small neural network predicts an input-dependent modulation of that scale from the same transformed features received by the GP kernel. On deterministic simulation data, this term is interpreted as model-discrepancy variance rather than physical observation noise; on data with genuine conditional variability, it can retain its conventional aleatoric interpretation. The total predictive variance is the sum of the epistemic and residual components.

\begin{figure}[t]
\centering
\includegraphics[width=\textwidth]{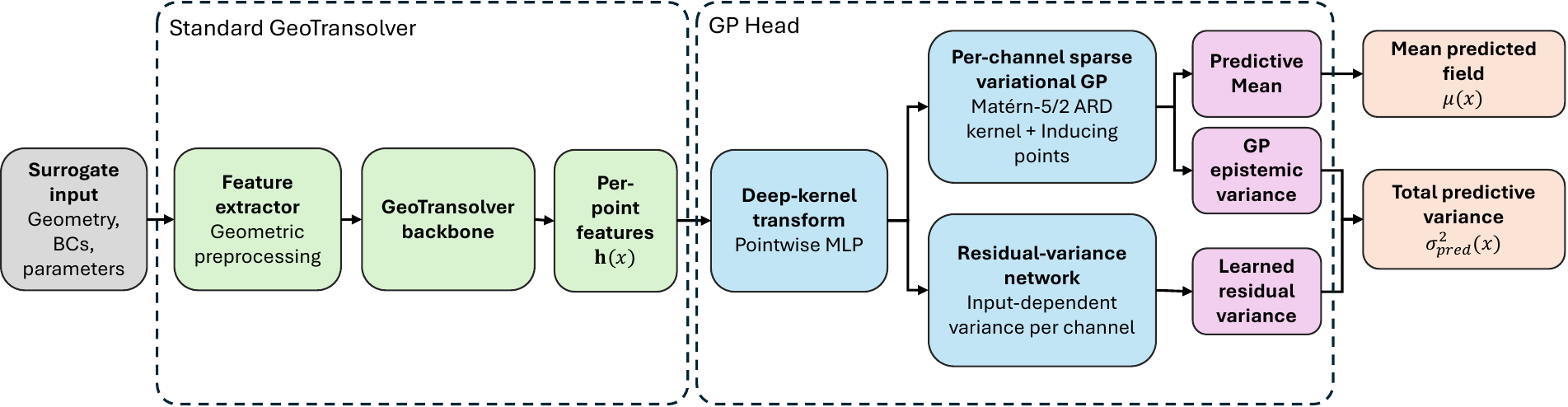}
\caption{Data flow of the evaluated GP-based surrogate. Per-point backbone features are transformed before entering independent inducing-point variational GPs and the residual-variance network. The GP epistemic and residual variances are summed to obtain total predictive variance.}
\label{fig:gp-schematic}
\end{figure}

Figure~\ref{fig:gp-schematic} summarizes the evaluated GP-based implementation. 

A single inference pass returns the predicted field, total variance, epistemic variance, and corresponding Gaussian prediction intervals. The total variance is used when evaluating interval coverage and calibration, while the epistemic component is the relevant quantity for model diagnostics and active-learning acquisition. Case-level OOD results report GP total and epistemic standard deviations separately: total variance describes the complete predictive estimate, while GP epistemic variance is used when comparing the GP with the epistemic disagreement estimated by the sampling methods.

\subsection{Concrete MC dropout}

Concrete MC dropout estimates model uncertainty by evaluating the same trained network multiple times with stochastic dropout retained during inference \citep{gal2016dropout}. Each pass produces a different field prediction, and the mean and variance across these predictions form an empirical predictive distribution. 

Different layers of a geometry-conditioned surrogate process distinct representations and can contribute differently to predictive uncertainty. This motivates adapting the strength of stochastic masking across the network. A single shared dropout probability restricts this flexibility. Conventional MC dropout can use separate layerwise rates, but selecting them through ablation requires a high-dimensional hyperparameter search whose cost grows with the number of dropout sites. Concrete dropout instead uses a continuous relaxation of the Bernoulli mask, allowing each probability $p_l$ to be learned jointly with the model parameters \citep{gal2017concrete}. The resulting probabilities provide adaptive, layer-specific stochastic masking; they do not directly measure the predictive uncertainty attributable to each layer.

The spread across stochastic passes estimates predictive disagreement induced by stochastic masking. It does not explicitly measure distance from the training data and is therefore not guaranteed to increase for unfamiliar inputs. In the implementations evaluated here, the entropy-regularized regression objective has the form
\begin{equation}
\mathcal{L}_{\mathrm{CD}}
=\frac{1}{N}\sum_{i=1}^{N}
\lVert y_i-f_{\mathbf{W},\tilde{\mathbf{z}}}(x_i)\rVert_2^2
+\lambda_{\mathrm{reg}}\sum_l
\left[p_l\log p_l+(1-p_l)\log(1-p_l)\right],
\label{eq:concrete-objective}
\end{equation}
where $\tilde{\mathbf{z}}$ denotes a reparameterized concrete mask.

The placement of concrete dropout layers varies with the backbone architecture. Across the evaluated configurations, we apply them at available sites within geometry/context projection and transformer processing, including context-projector outputs, attention-output projections, and post-attention or post-feed-forward residual updates. Input-feature construction and preprocessing, as well as the final projection to physical output variables, remain deterministic.

For layer $l$, the implemented mask draws $u\sim\mathcal{U}(0,1)$ and computes
\begin{equation}
d_l=\operatorname{sigmoid}\!\left(
\frac{\operatorname{logit}(u)+\operatorname{logit}(p_l)}{\tau}
\right),
\qquad
\tilde{\mathbf{z}}_l=\frac{1-d_l}{1-p_l+\epsilon},
\end{equation}
where $d_l$ is the relaxed drop indicator and $\tau=0.1$. All evaluated models initialize $p_l=0.05$. Initialization is clipped to $[0.01,0.99]$ for numerical safety, and the learned logit is constrained to $[-10,10]$ during stochastic evaluation, corresponding approximately to $p_l\in[4.5\times10^{-5},0.99995]$. The reported entropy coefficient is $\lambda_{\mathrm{reg}}=10^{-4}$ in Equation~\ref{eq:concrete-objective}; because the bracketed term is negative Bernoulli entropy, minimizing the objective discourages probabilities from collapsing to zero or one.

Concrete MC dropout requires one training run and stores one checkpoint. At inference, the concrete dropout layers remain stochastic while the rest of the model is in evaluation mode. For $S$ passes, the reported moments are
\begin{equation}
\bar{\mathbf{y}}=\frac{1}{S}\sum_{s=1}^{S}\mathbf{y}^{(s)},\qquad
\widehat{\mathrm{Var}}(\mathbf{y})=\frac{1}{S-1}\sum_{s=1}^{S}
\left(\mathbf{y}^{(s)}-\bar{\mathbf{y}}\right)^2 .
\end{equation}
Each pass also produces a complete field that can be propagated sample-wise to a derived quantity.

The implementation evaluated in this work reports only the spread across stochastic passes and therefore estimates epistemic disagreement rather than total predictive variance. Its scale is not fitted directly against observed residuals, and it may require calibration on held-out data before being interpreted as a predictive interval. A separate variance output could be added to model aleatoric variability or model discrepancy, but such an output is not used here.

\subsection{Deep Ensembles}

A deep ensemble constructs a predictive distribution from independently trained instances of the same model architecture \citep{lakshminarayanan2017ensembles}. The members differ through their random initialization and stochastic data ordering and may converge to different solutions of the training objective. At inference, each member predicts a complete field. The ensemble mean is used as the final prediction, while the spread across members estimates uncertainty arising from disagreement between the learned models.

Like MC dropout, ensemble disagreement contains no explicit comparison with the training inputs. Its usefulness for detecting unfamiliar inputs depends on whether the members respond differently in those regions, which must be verified empirically. Diversity is therefore central to the method: members that converge to highly correlated solutions produce narrow uncertainty estimates even if their shared prediction is inaccurate. The number of members also affects the stability of the estimated variance, particularly for the relatively small ensembles that are practical for large CAE surrogates.

The ensemble evaluated here contains $K$ independently initialized and trained models, stores $K$ checkpoints, and evaluates every member once per case. The members can be trained and evaluated concurrently, reducing wall-clock time but not total computation or storage.

The ensemble implementation evaluated here estimates epistemic disagreement only. It does not include an additional output for aleatoric or residual variance, although probabilistic ensemble members could be trained to provide one. Its spread should therefore not be interpreted automatically as total predictive uncertainty or a calibrated error interval.

\section{Experimental Design}

\label{sec:experimental-design}
We evaluate the UQ methods on three datasets spanning external aerodynamics and transient structural dynamics. DrivAerStar provides the complete comparison among the GP-based method, concrete MC dropout, and a deep ensemble. AirFRANS and the automotive crash study evaluate the GP-based and concrete MC dropout methods under a different distribution-shift axis and a different physical problem, respectively.

\subsection{Datasets and Distribution Shift Definitions}

DrivAerStar \citep{qiu2025drivaerstar} is a three-dimensional external-aerodynamics dataset containing vehicle geometries from three rear-end body-style classes. To enable testing across different body-styles (geometry distributions), all models are trained only on the Fastback class and evaluated on others. Fastback evaluation cases are therefore treated as in distribution, while Notchback and Estateback are held-out body-style classes. These classes provide a categorical geometry shift rather than a calibrated continuous distance from the training distribution. We evaluate 100 geometries from each class, for a total of 300 scored cases. The predicted fields are surface pressure and the three components of wall shear stress. Derived aerodynamic quantities, including drag and lift, are computed from these fields.

AirFRANS \citep{bonnet2022airfrans} contains two-dimensional Reynolds-averaged Navier-Stokes simulations over airfoil geometries. The prediction targets are the two velocity components, pressure, and turbulent viscosity. The fitting split contains 645 cases with angles of attack between $-2.48^\circ$ and $12.45^\circ$, and 159 additional cases from the same range are used for in-distribution validation. The out-of-distribution split contains 196 cases whose angles of attack extend outside the fitting range at both ends, i.e. from $-4.94^\circ$ to $-2.48^\circ$ and $12.45^\circ$ to $14.93^\circ$. Unlike DrivAerStar, the unfamiliarity axis is a boundary condition rather than a held-out geometry class. This gives a continuous extrapolation test under such distribution shift. A $\log(1+x)$ transformation is applied to turbulent viscosity for all reported methods, and normalization statistics are computed from the fitting split only.

The automotive crash dataset is based on the body-in-white finite-element study introduced by \citet{nabian2026automotivecrash}. It contains 150 transient structural simulations on a common mesh with 384,862 nodes. The design variables are the thicknesses of 33 front-end structural components, varied within ±20\% of their nominal values. In the curated dataset, component thickness is represented as a static per-node input on the common mesh. The targets are three-dimensional nodal displacements over 26 time steps spanning 0--125 ms. We use 135 runs for training and 15 for evaluation, with 8 runs monitored as validation during training and 7 retained as genuinely held-out cases. However, the crash experiments are not explicit out-of-distribution tests. The held-out designs lie within the sampled design space, and the most difficult responses are not necessarily unusual according to input-space distance. Instead, this dataset examines whether uncertainty remains informative across designs, spatial locations, and time in a strongly nonlinear transient problem.

Table~\ref{tab:datasets} summarizes the three datasets, prediction tasks, and evaluation splits.
\begin{table}[t]
\centering
\small
\begin{tabular}{p{0.15\textwidth}p{0.21\textwidth}p{0.24\textwidth}p{0.27\textwidth}}
\toprule
\textbf{Dataset} & \textbf{Prediction task} & \textbf{Training and validation} & \textbf{Evaluation setting} \\
\midrule
DrivAerStar & Surface pressure and three wall-shear-stress components & 3,252 Fastback training geometries and 813 Fastback validation geometries, with 100 validation geometries reserved for calibration & Disjoint test sets containing 100 Fastback, 100 Notchback, and 100 Estateback geometries \\
AirFRANS & $U_x$, $U_y$, pressure, and turbulent viscosity & 645 fitting cases with $\alpha \in [-2.48^\circ,12.45^\circ]$ and 159 in-distribution validation cases & 196 extrapolation cases with
$\alpha\in[-4.94^\circ,-2.48^\circ)\cup(12.45^\circ,14.93^\circ]$\\
Automotive crash & Three-dimensional nodal displacement over time & 135 training runs from a 150-run design study & 8 monitored validation runs and 7 held-out runs over 26 time steps from 0 to 125 ms \\
\bottomrule
\end{tabular}
\caption{Datasets and evaluation settings. Each dataset examines a different source of difficulty: categorical geometry shift, continuous boundary condition extrapolation, or transient response variability.}
\label{tab:datasets}
\end{table}

Table~\ref{tab:comparison-matrix} identifies the methods evaluated on each dataset.

\begin{table}[t]
\centering
\small
\begin{tabular}{lcccc}
\toprule
\textbf{Dataset} & \textbf{Deterministic} & \textbf{GP-based} &
\textbf{Concrete MC dropout} & \textbf{Ensemble} \\
\midrule
DrivAerStar      & \cmark & \cmark & \cmark & \cmark \\
AirFRANS         & \cmark & \cmark & \cmark & \xmark \\
Automotive crash & \cmark & \cmark & \cmark & \xmark \\
\bottomrule
\end{tabular}
\caption{Methods evaluated on each dataset. A check mark denotes an included method; a cross denotes a method that was not evaluated.}
\label{tab:comparison-matrix}
\end{table}

For DrivAerStar, all uncertainty methods use the same GeoTransolver backbone. The model receives the vehicle surface mesh and its geometric features, including point coordinates and surface normals, together with the freestream velocity and air density. Because the boundary conditions are fixed across all cases in this study, geometry is the only varying conditioning input. The model predicts surface pressure and the three wall-shear-stress components at each mesh point. The deep ensemble contains five independently trained members, and its first member also serves as the deterministic reference.

AirFRANS uses the same GeoTransolver architecture across its deterministic, GP-based, and concrete MC dropout configurations. The model receives the airfoil geometry, the surrounding volume-mesh coordinates, the angle of attack $\alpha$, and the freestream-velocity magnitude $U_\infty$. It predicts the two in-plane velocity components, pressure, and turbulent viscosity throughout the flow domain. In contrast to DrivAerStar, both geometry and boundary conditions vary between cases, and the out-of-distribution split is defined by angles of attack outside the range used for training.

The automotive crash study instead uses the time-conditional GeoTransolver with FLARE attention configuration, which is held fixed across the methods evaluated on that dataset. This backbone was selected based on prior results on the same full-vehicle crash benchmark, where the FLARE-based modification to GeoTransolver achieved lower relative \(L_2\) error and approximately halved peak attention-block memory \citep{akhare2026highfidelity}. The model receives the undeformed reference geometry, component thicknesses, and normalized query time and predicts the nodal displacement at that time. During training, one target time step is sampled per example; during inference, the model is evaluated at each of the 25 non-initial time steps to construct the complete trajectory.

\subsection{Metrics}
\label{sec:metrics}
We use complementary metrics to evaluate predictive accuracy, interval magnitude, error discrimination, response to distribution shift, and uncertainty in derived engineering quantities. These metrics answer different questions and should not be interpreted as interchangeable measures of overall UQ quality.

Let $y_i$ denote a target value, $\mu_i$ the predictive mean, $\sigma_i$ the predicted standard deviation, and $e_i=y_i-\mu_i$ the prediction error at a point, time, and output channel. When a method provides both total and epistemic variance, the metrics are computed for each component where relevant. Table~\ref{tab:metric-map} maps the reported metrics to the uncertainty property they evaluate.
\begin{table}[H]
\centering
\small
\begin{tabular}{p{0.24\textwidth}p{0.28\textwidth}p{0.38\textwidth}}
\toprule
\textbf{Evaluation question} & \textbf{Metrics} & \textbf{What they measure} \\
\midrule
Is the predictive mean accurate? & Relative $L_2$ error and RMSE & Error in the mean prediction, independent of uncertainty quality \\
Are interval magnitudes credible? & $z$-RMS, coverage, NLPD, and sharpness & Agreement between predicted interval width and observed residuals \\
Does uncertainty identify high error? & Spearman rank correlation and AUSE & Agreement between uncertainty ordering and error ordering \\
Does uncertainty respond to unfamiliar inputs? & OOD AUROC and growth ratio & Case-level separation and whether uncertainty increases with error under shift \\
Is uncertainty informative for an engineering quantity? & Calibration and error--uncertainty correlation for the derived quantity & Whether field uncertainty remains useful after propagation to the reported decision variable \\
\bottomrule
\end{tabular}
\caption{Evaluation questions and their corresponding metrics. No single metric evaluates all downstream uses of uncertainty.}
\label{tab:metric-map}
\end{table}
Predictive accuracy is measured using relative $L_2$ error,
\begin{equation}
\epsilon_{L_2}=\frac{\lVert \mu-y\rVert_2}{\lVert y\rVert_2},
\end{equation}
together with root-mean-square error where an absolute error in physical units is required. Magnitude credibility is evaluated using the standardized residual $z_i=e_i/\sigma_i$. We report
\begin{equation}
z\text{-RMS}=\sqrt{\frac{1}{N}\sum_{i=1}^{N}z_i^2}.
\end{equation}
A value of one means that the standardized residuals have unit RMS across the evaluated observations. Values greater than one indicate that errors are larger than predicted, while values below one indicate intervals that are broader than required. This does not require $|e_i|=\sigma_i$ at every observation; individual standardized residuals can vary.
The $95\%$ pointwise coverage is
\begin{equation}
\mathrm{Coverage}_{95}=\frac{1}{N}\sum_{i=1}^{N}\mathbb{I}\left(|e_i|\leq1.96\sigma_i\right).
\end{equation}
Its nominal target is $0.95$ for a calibrated Gaussian predictive distribution. This is pointwise coverage and should not be interpreted as simultaneous coverage of an entire field.
Negative log predictive density evaluates the predictive mean and variance jointly:
\begin{equation}
\mathrm{NLPD}=\frac{1}{N}\sum_{i=1}^{N}\left[\frac{1}{2}\log\left(2\pi\sigma_i^2\right)+\frac{e_i^2}{2\sigma_i^2}\right].
\end{equation}
Lower values indicate a better predictive distribution. Unlike coverage, NLPD penalizes both intervals that are too narrow and intervals that are unnecessarily broad. Sharpness is the mean predicted standard deviation,
\begin{equation}
\mathrm{Sharpness}=\frac{1}{N}\sum_{i=1}^{N}\sigma_i.
\end{equation}
Sharpness is reported in the physical units of the target and indicates whether an interval is narrow enough to be operationally useful. It is not a measure of quality by itself: smaller uncertainty is desirable only when calibration and predictive accuracy remain comparable.

Error discrimination is evaluated using Spearman rank correlation between $|e_i|$ and $\sigma_i$. A high positive value indicates that locations or cases assigned greater uncertainty also tend to have larger errors. Rank correlation is invariant to a positive global rescaling of uncertainty and therefore evaluates a different property from calibration. 

We also report the area under the sparsification error curve (AUSE) \citep{ilg2018uncertainty}. In the DrivAerStar implementation, each case contributes one RMS error and one mean standard deviation per output channel. Cases are removed in decreasing order of uncertainty, and the remaining RMSE is compared with an oracle that removes the largest realized errors first. Each curve is normalized by its initial RMSE. AUSE is the area between the two curves, averaged over channels; zero represents perfect case ordering.

OOD detection is evaluated at the case level using the area under the receiver operating characteristic curve (AUROC). Each case is assigned a scalar score equal to its pointwise mean predicted standard deviation over the evaluated field. GP total and GP epistemic scores are reported separately. The score is used to distinguish an out-of-distribution split from the corresponding in-distribution split. An AUROC of $0.5$ indicates chance-level separation, while a value of one indicates perfect separation.
OOD AUROC measures case separation but not whether uncertainty changes in proportion to the accompanying change in error. To examine this second property, we define the following growth ratio for this study:
\begin{equation}
\mathrm{GR}=\frac{\sigma_{\mathrm{ood}}/\sigma_{\mathrm{id}}}{\mathrm{RMSE}_{\mathrm{ood}}/\mathrm{RMSE}_{\mathrm{id}}}.
\end{equation}

A value of one indicates matched relative changes in uncertainty and RMSE. A value below one means that the uncertainty ratio is smaller than the error ratio; when error increases under shift, uncertainty does not increase by a matching proportion. This diagnostic is invariant to a fixed positive rescaling applied to both splits and is not itself a calibration guarantee.

We estimate drag uncertainty in two ways. First, for all three methods, we propagate the pointwise marginal variances of pressure and wall shear stress through the drag integral, assuming independence across surface points and output channels. This provides a common basis for comparison, but neglects spatial and cross-channel covariance. Second, for concrete MC dropout and the ensemble, we compute drag directly from each sampled field and use the resulting variation to estimate its uncertainty. Because each field is propagated as a whole, this calculation preserves the dependencies represented in the sampled predictions. Equation~\ref{eq:drag-variance-propagation} defines the marginal approximation.

\paragraph{Metric aggregation policy}
Unless stated otherwise in a table, pointwise metrics are first reduced over the valid mesh points within each case and output channel. They are then averaged over channels and cases, with each case receiving equal weight. Relative $L_2$ error is computed separately for each case and channel before averaging. Spearman correlation is computed between pointwise absolute error and predicted standard deviation within each case and channel, and the resulting correlations are then averaged. Crash mesh metrics pool the three displacement channels over the evaluated nodes and 25 non-initial time steps. Probe metrics pool the specified probe nodes and held-out runs at each time step before aggregation over time. AUSE follows the case-level construction defined above. DrivAerStar growth ratios compare pressure RMSE with pressure standard deviation within each body style. For AirFRANS, the growth ratio is computed separately for each of the four output channels and then averaged across channels.

Reported intervals quantify sensitivity to the finite test set using case-level bootstrap resampling. For each bootstrap replicate, we sample cases with replacement while preserving the original test-set size and recompute the metric. The central 95\% of the resulting values defines the reported interval. When comparing methods, the same resampled cases are used for every method.

\subsection{Post-Hoc Calibration Protocol}

We report raw and post-hoc calibrated uncertainty separately. Calibration rescales the predicted standard deviation without changing the predictive mean. For calibration group $g$, let $\mathcal{I}_g$ denote the set of observations assigned to that group and let $n_g=|\mathcal{I}_g|$. The scale factor is

\begin{equation}
s_g=
\sqrt{
\frac{1}{n_g}
\sum_{i\in\mathcal{I}_g}
\frac{e_i^2}{\sigma_i^2}
},
\end{equation}

and the calibrated standard deviation is $\sigma'_i=s_{g(i)}\sigma_i$, where $g(i)$ denotes the calibration group associated with observation $i$. The groups correspond to output channels for DrivAerStar and AirFRANS and to time steps for crash. By construction, the fitted scale gives $z$-RMS equal to one within each calibration group on the calibration data. The values reported on disjoint evaluation cases therefore measure how well that fitted scale transfers.

The calibration and evaluation splits are summarized in Table~\ref{tab:datasets}. For DrivAerStar, the scale factors are fitted on a reserved subset of 100 Fastback validation geometries that is disjoint from all reported test cases. The factors are then applied unchanged to the Fastback, Notchback, and Estateback test sets, allowing us to distinguish a correctable in-distribution scale mismatch from a failure of calibration to transfer under geometry shift.

For AirFRANS, the scale factors are fitted on 159 in-distribution validation cases whose angles of attack lie within the fitting range. These cases are disjoint from both the 645 model-fitting cases and the 196 extrapolation cases. The fitted factors are applied unchanged to the extrapolation cases, testing whether an in-distribution calibration transfers beyond the angle-of-attack range used for fitting. For crash, the time-dependent factors are fitted on the eight monitored validation runs and applied unchanged to the seven held-out runs.

\subsection{Implementation and Reproducibility}
All three UQ approaches are implemented within the PhysicsNeMo framework \citep{physicsnemo}. The DrivAerStar evaluation is implemented through the method-agnostic UQ benchmarking framework in PhysicsNeMo-CFD \citep{physicsnemocfd}. The UQ benchmarking capability extends the deterministic benchmarking framework introduced by \citet{tangsali2025benchmarking}, which standardized the evaluation of predictive accuracy, computational performance, scalability, and generalization for automotive-aerodynamics surrogates. The extension represents each prediction through a common distribution interface containing its mean, total uncertainty, epistemic uncertainty, and samples when available, allowing the same cases, deterministic accuracy metrics, and UQ metrics to be applied consistently to closed-form and sampling-based predictions.

Within each dataset, the available methods use the same training data, normalization, and evaluation cases. The crash configurations share this backbone but use method-specific point-sampling budgets during training. All reported training runs use eight GPUs. DrivAerStar and AirFRANS use the final checkpoint at epoch 100. In crash, we train both methods till 200 epochs and then select the best checkpoint based on validation accuracy. Concrete MC dropout uses epoch 30 and the GP uses epoch 160, selected by the lowest aggregate validation displacement relative $L_2$ error; the deterministic reference uses its final retained checkpoint at epoch 200. \ref{app:configurations} records the sampling sizes, model size, and other details.

For DrivAerStar, we also trained the GP-based and concrete MC dropout configurations using three random seeds per method to assess sensitivity to training initialization. Run-to-run variability was small and did not alter the principal method comparisons, so the detailed results report one training run for each method. ~\ref{app:seed-sensitivity} provides the complete sensitivity analysis.

\section{Controlled Comparison on DrivAerStar}
\label{sec:drivaerstar-results}

\begin{figure}[t]
\centering
\includegraphics[width=\textwidth]{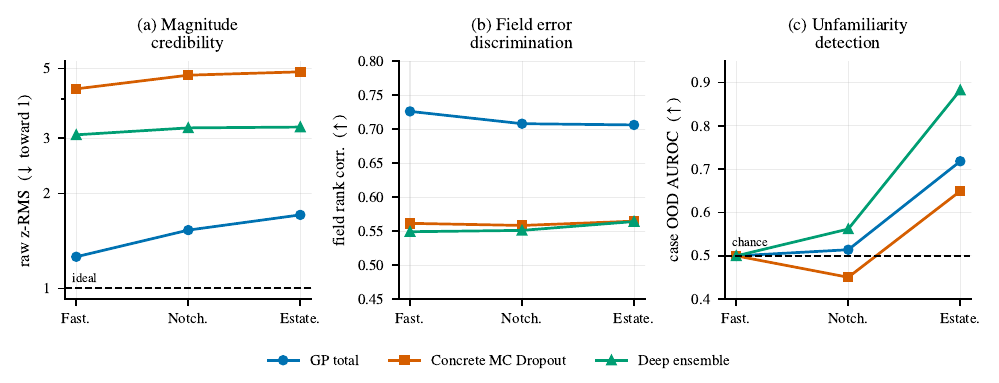}
\caption{The DrivAerStar comparison viewed through three uncertainty properties. Panel (a) reports raw $z$-RMS, for which values closer to one (dotted line) indicate a more credible aggregate uncertainty scale; values above one indicate intervals that are too narrow, whereas values below one indicate intervals that are broader than required. Panel (b) reports local field-error discrimination using Spearman correlation, for which higher values indicate stronger correspondence between predicted uncertainty and absolute error. Panel (c) reports case-level OOD detection using AUROC, for which higher values indicate better separation from Fastback and an AUROC of $0.5$ indicates no better separation than random ordering. The GP curve uses total predictive standard deviation, while the sampling curves use concrete MC dropout or ensemble disagreement.}
\label{fig:drivaer-properties}
\end{figure}

DrivAerStar provides the complete controlled comparison among the GP-based method, concrete MC dropout, and the five-member deep ensemble. We organize the results according to the downstream question being asked of the uncertainty: predictive accuracy, interval magnitude, error discrimination, response to geometric shift, and informativeness for an engineering quantity. Figure~\ref{fig:drivaer-properties} provides an overview, and computational requirements are considered separately at the end of the section. The detailed tables and figures report one trained GP-based model, one trained concrete MC dropout model, and the five-member deep ensemble.

\subsection{Predictive Accuracy}

We first examine mean-prediction errors to determine whether adding a UQ mechanism materially changes the underlying surrogate. Table~\ref{tab:drivaer-accuracy} reports relative $L_2$ errors for the predicted fields and relative drag-coefficient error. The deterministic reference is the first member of the deep ensemble, so its comparison with the ensemble isolates the effect of averaging the five member predictions.

\begin{table}[!htbp]
\centering
\small
\begin{tabular}{llccccc}
\toprule
\textbf{Method} & \textbf{Class} & \textbf{Pressure} & \textbf{WSS-$x$} & \textbf{WSS-$y$} & \textbf{WSS-$z$} & \textbf{Drag} \\
\midrule
Deterministic & F & 0.231 & 0.235 & 0.380 & 0.220 & 0.044 \\
              & N & 0.235 & 0.240 & 0.394 & 0.224 & 0.040 \\
              & E & 0.304 & 0.264 & 0.400 & 0.231 & 0.097 \\
\midrule
GP-based     & F & 0.225 & 0.228 & 0.379 & 0.217 & 0.056 \\
             & N & 0.232 & 0.238 & 0.395 & 0.237 & 0.046 \\
             & E & 0.278 & 0.248 & 0.404 & 0.254 & 0.071 \\
\midrule
Concrete MC dropout & F & 0.205 & 0.216 & 0.354 & 0.207 & 0.060 \\
             & N & 0.212 & 0.225 & 0.373 & 0.221 & 0.053 \\
             & E & 0.244 & 0.231 & 0.376 & 0.217 & 0.063 \\
\midrule
Deep ensemble & F & 0.208 & 0.218 & 0.353 & 0.203 & 0.057 \\
              & N & 0.213 & 0.223 & 0.367 & 0.210 & 0.043 \\
              & E & 0.280 & 0.247 & 0.372 & 0.214 & 0.099 \\
\bottomrule
\end{tabular}
\caption{Mean-prediction errors on Fastback (F), Notchback (N), and Estateback (E). Field entries are relative $L_2$ errors and drag is relative coefficient error. Lift is omitted because its reference value approaches zero for some Estateback cases, making relative error unstable.}
\label{tab:drivaer-accuracy}
\end{table}

Across the 36 matched field-and-class comparisons, the absolute difference in relative $L_2$ error, normalized by the corresponding deterministic value, averages 6.4\%. The evaluated UQ models therefore show no systematic degradation in mean-prediction accuracy relative to the deterministic reference. Because the deterministic, GP-based, and concrete MC dropout models were trained separately, and the sampling-based methods average stochastic predictions or ensemble members, individual differences cannot be attributed solely to the UQ mechanism. No method is consistently the most accurate across both fields and derived quantities.

The accuracy results serve as a control: the subsequent UQ comparisons are not dominated by a severely degraded predictive mean. They also show why UQ metrics must be interpreted alongside accuracy. Broad intervals around an inaccurate mean, or apparently stable uncertainty when the mean already underfits, should not be interpreted as favorable UQ behavior.

\begin{figure}[!htbp]
\centering
\includegraphics[width=\textwidth]{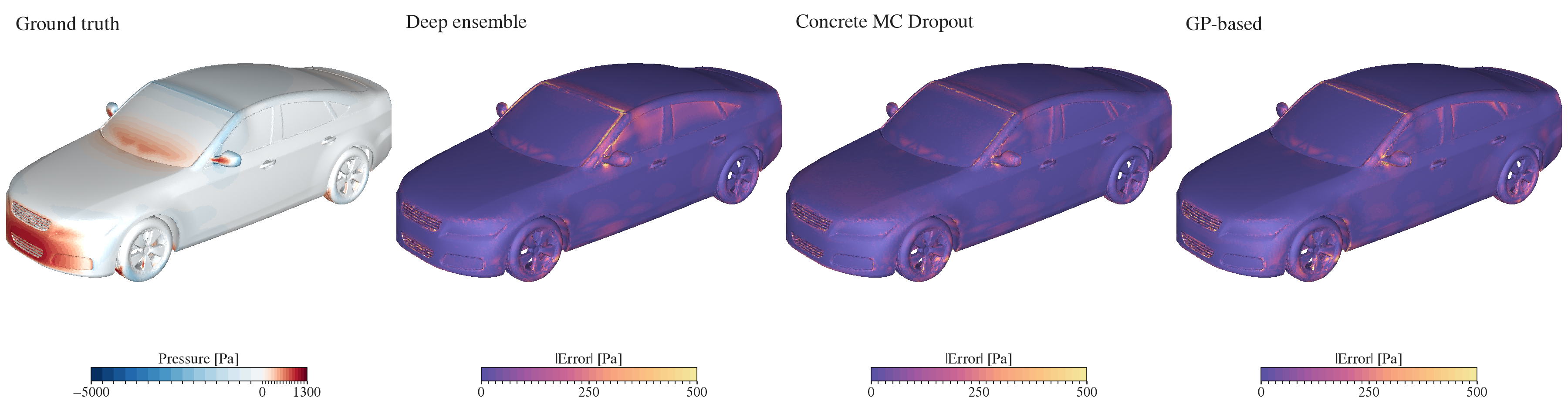}
\caption{Ground-truth surface pressure and absolute pressure error for the ensemble, concrete MC dropout, and GP-based predictive means on one representative Fastback case. All error panels use the same color scale.}
\label{fig:drivaer-fastback-error}
\end{figure}

Figure~\ref{fig:drivaer-fastback-error} provides a representative spatial view of the mean-prediction errors summarized in Table~\ref{tab:drivaer-accuracy}.

\FloatBarrier

\FloatBarrier
\subsection{Magnitude Credibility}

We next ask whether the numerical magnitude of the predicted uncertainty is consistent with the observed residuals. Table~\ref{tab:drivaer-raw-magnitude} reports raw results before any post-hoc recalibration.

\begin{table}[!htbp]
\centering
\small
\begin{tabular}{llccccc}
\toprule
\textbf{Method} & \textbf{Class} & \textbf{$z$-RMS} & \textbf{95\% cov.} & \textbf{NLPD} & \textbf{Total $\sigma$} & \textbf{Epistemic $\sigma$} \\
\midrule
GP total & F & 1.252 & 0.900 & 0.972 & 9.350 & 0.894 \\
         & N & 1.515 & 0.875 & 1.357 & 9.351 & 0.872 \\
         & E & 1.687 & 0.855 & 1.675 & 9.709 & 0.928 \\
\midrule
Concrete MC dropout & F & 4.274 & 0.511 & 8.539 & 3.088 & 3.088 \\
           & N & 4.732 & 0.488 & 10.631 & 3.049 & 3.049 \\
           & E & 4.847 & 0.482 & 11.256 & 3.177 & 3.177 \\
\midrule
Deep ensemble & F & 3.053 & 0.681 & 4.409 & 5.226 & 5.226 \\
              & N & 3.218 & 0.662 & 4.944 & 5.262 & 5.262 \\
              & E & 3.235 & 0.666 & 5.084 & 5.787 & 5.787 \\
\bottomrule
\end{tabular}
\caption{Raw interval-magnitude metrics. The target values are one for $z$-RMS and 0.95 for pointwise coverage. Sharpness is the mean predicted standard deviation in the physical units of the pressure and wall-shear-stress targets. For the sampling methods, total and epistemic uncertainty coincide because no separate residual-variance output is used.}
\label{tab:drivaer-raw-magnitude}
\end{table}

Without post-hoc calibration, the evaluated GP total variance is closest to the nominal predictive interval scale. Its Fastback $z$-RMS is 1.252 with 90.0\% coverage, compared with $z$-RMS values of 4.274 and 3.053 for concrete MC dropout and the ensemble. The intervals remain narrower than required for nominal coverage.

The GP epistemic standard deviation is approximately 9.6\% of its total standard deviation. The raw interval result is therefore primarily associated with the learned residual-variance model, not the distance-aware posterior variance; see Section~\ref{sec:variance-decomposition}.

Table~\ref{tab:drivaer-calibrated-magnitude} reports results after applying the per-channel scales fitted on the independent Fastback calibration subset. The fitted standard-deviation multipliers range from 1.19 to 1.40 for the GP-based method, 3.62 to 5.01 for concrete MC dropout, and 2.61 to 3.44 for the ensemble.

\begin{table}[!htbp]
\centering
\small
\begin{tabular}{llccc}
\toprule
\textbf{Method} & \textbf{Class} & \textbf{Calibrated $z$-RMS} & \textbf{Calibrated 95\% cov.} & \textbf{Calibrated NLPD} \\
\midrule
GP total & F & 1.003 & 0.949 & 0.907 \\
         & N & 1.209 & 0.927 & 1.144 \\
         & E & 1.342 & 0.911 & 1.337 \\
\midrule
Concrete MC dropout & F & 0.999 & 0.948 & 1.242 \\
           & N & 1.106 & 0.933 & 1.346 \\
           & E & 1.134 & 0.929 & 1.416 \\
\midrule
Deep ensemble & F & 0.994 & 0.954 & 1.321 \\
              & N & 1.047 & 0.947 & 1.383 \\
              & E & 1.055 & 0.947 & 1.473 \\
\bottomrule
\end{tabular}
\caption{Interval-magnitude metrics after fitting one standard-deviation scale per output channel on 100 independent Fastback calibration geometries. The factors are applied unchanged to the disjoint Fastback, Notchback, and Estateback test cases.}
\label{tab:drivaer-calibrated-magnitude}
\end{table}

All three methods reach approximately nominal magnitude and coverage on the disjoint Fastback test set. The large raw $z$-RMS values of the sampling methods therefore arise primarily from a correctable in-distribution scale mismatch rather than an absence of useful variation in their uncertainty fields. Nevertheless, the amount of correction required remains operationally relevant: without calibration data, the raw sampling spreads would produce intervals that are substantially too narrow.

Calibration transfer under geometry shift produces a different ordering. The ensemble remains close to nominal on both Notchback and Estateback, while concrete MC dropout becomes moderately overconfident and the GP-based method degrades most. At the same time, the GP retains the lowest NLPD on every class. NLPD evaluates mean accuracy and sharpness in addition to standardized residual scale, whereas $z$-RMS and coverage focus more directly on interval magnitude. Thus, even within the broad category of calibration metrics, the relative performance depends on the criterion being used.

\FloatBarrier
\subsection{Error Discrimination}

Magnitude credibility does not establish whether uncertainty identifies inaccurate predictions. This distinction is operationally relevant as a reliable error ordering can help prioritize locations or cases for review and solver verification and, when based on epistemic uncertainty, inform the acquisition of additional training data. We evaluate local ordering with within-case pointwise Spearman correlation and case ordering with AUSE. Both depend on uncertainty ranking rather than absolute scale, so per-channel post-hoc rescaling does not alter the corresponding within-channel rankings.

Table~\ref{tab:drivaer-discrimination} reports the local and case-level error-discrimination metrics for each uncertainty estimate. 

\begin{table}[!htbp]
\centering
\small
\begin{tabular}{llcc}
\toprule
\textbf{Method} & \textbf{Class} & \textbf{Spearman $\rho$} & \textbf{AUSE} \\
\midrule
GP total & F & 0.726 & 0.0218 \\
         & N & 0.708 & 0.0234 \\
         & E & 0.706 & 0.0332 \\
\midrule
GP epistemic & F & 0.612 & 0.0295 \\
             & N & 0.607 & 0.0235 \\
             & E & 0.614 & 0.0323 \\
\midrule
Concrete MC dropout & F & 0.561 & 0.0413 \\
           & N & 0.558 & 0.0364 \\
           & E & 0.565 & 0.0302 \\
\midrule
Deep ensemble & F & 0.549 & 0.0501 \\
              & N & 0.551 & 0.0336 \\
              & E & 0.564 & 0.0331 \\
\bottomrule
\end{tabular}
\caption{Error discrimination. Spearman measures pointwise ordering within cases; AUSE sparsifies complete cases using channel RMS error and mean standard deviation. Higher Spearman and lower AUSE are better.}
\label{tab:drivaer-discrimination}
\end{table}

All three methods contain meaningful local error information. Their rank correlations range from approximately 0.55 to 0.73. The GP total uncertainty has the highest correlation on each body style, while its epistemic component remains positively associated with error but is less discriminative than the total variance. The sampling methods also rank errors usefully despite their poor raw interval magnitude. Dropout, for example, has a Fastback $z$-RMS of 4.274 but a rank correlation of 0.561. Its raw uncertainty scale is unsuitable as a predictive interval without recalibration, yet its ordering can still support tasks such as identifying high-error regions or prioritizing locations for review.  

\begin{figure}[!htbp]
\centering
\includegraphics[width=\textwidth]{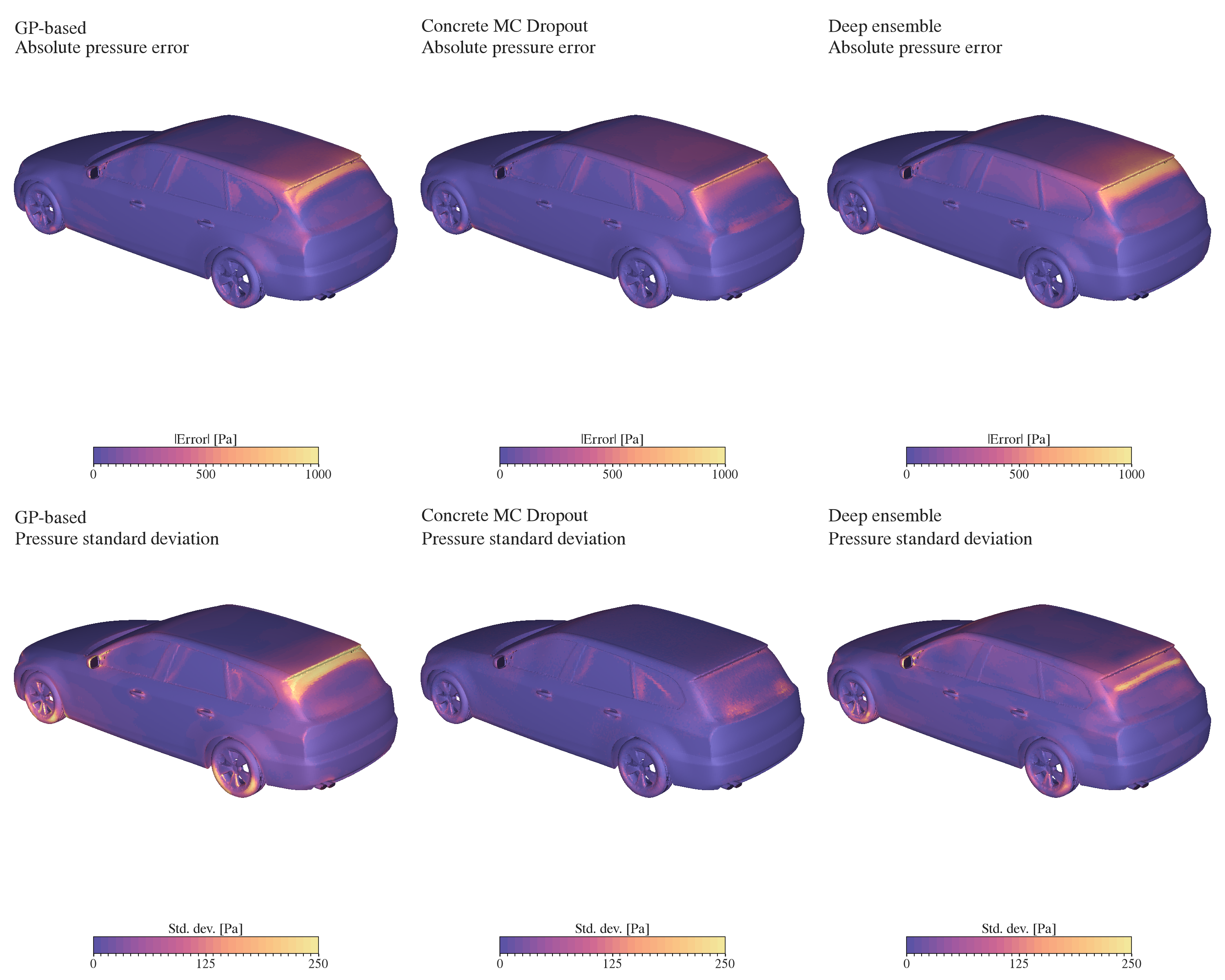}
\caption{Absolute pressure error (top) and raw predicted pressure standard deviation (bottom) on one Estateback geometry. Columns show GP total, concrete MC dropout disagreement, and deep-ensemble disagreement.}
\label{fig:drivaer-estateback-fields}
\end{figure}

Figure~\ref{fig:drivaer-estateback-fields} should be read columnwise: useful local uncertainty should be larger in regions with larger absolute pressure error. Across all three methods, elevated uncertainty aligns with the dominant error region around the roof trailing edge and adjacent rear pillar. The methods differ in the secondary regions emphasized around the wheels and body side. This recurring spatial correspondence between high-error and high-uncertainty regions provides a spatial interpretation of the positive Spearman correlations reported in Table~\ref{tab:drivaer-discrimination}. Because the error and uncertainty rows use different numerical ranges, the figure illustrates spatial error ordering rather than calibration of uncertainty magnitude.

\FloatBarrier

\FloatBarrier
\subsection{Response to Geometric Shift}

We next evaluate whether uncertainty responds when the body style changes from the Fastback class used for training to the held-out Notchback and Estateback classes. Two aspects are considered. The growth ratio compares the relative change in uncertainty with the relative change in error, while case-level AUROC measures whether aggregate uncertainty can distinguish unfamiliar geometries from Fastbacks.

Table~\ref{tab:drivaer-shift} reports pressure-channel growth, using pressure standard deviation and pressure RMSE in the same physical units. Values are normalized by the corresponding Fastback quantities. The AUROC score uses each case's mean field uncertainty.

\begin{table}[!htbp]
\centering
\scriptsize
\begin{tabular}{llcccc}
\toprule
\textbf{Method} & \textbf{Class} & \textbf{Pressure $\sigma$ growth} &
\textbf{Pressure RMSE growth} &
\textbf{Growth ratio} & \textbf{AUROC [95\% CI]} \\
\midrule
GP total & N & 1.001 & 1.045 & 0.957 & 0.514 [0.432, 0.595] \\
         & E & 1.039 & 1.235 & 0.841 & 0.719 [0.646, 0.788] \\
\midrule
GP epistemic & N & 0.977 & 1.045 & 0.934 & 0.452 [0.372, 0.533] \\
             & E & 1.038 & 1.235 & 0.840 & 0.684 [0.605, 0.759] \\
\midrule
Concrete MC dropout & N & 0.988 & 1.049 & 0.941 & 0.451 [0.370, 0.531] \\
           & E & 1.029 & 1.183 & 0.869 & 0.650 [0.571, 0.725] \\
\midrule
Deep ensemble & N & 1.007 & 1.037 & 0.971 & 0.562 [0.480, 0.641] \\
              & E & 1.108 & 1.346 & 0.823 & 0.883 [0.832, 0.928] \\
\bottomrule
\end{tabular}
\caption{Response to Notchback (N) and Estateback (E) relative to Fastback. Pressure $\sigma$ and RMSE growth are measured relative to Fastback; a growth ratio of one indicates matched growth. AUROC uses the all-channel case score (higher is better; 0.5 is chance). Brackets show 95\% bootstrap confidence intervals (CIs).}
\label{tab:drivaer-shift}
\end{table}


For all three methods, the relative change in predicted uncertainty is smaller than the relative change in prediction error.  Relative to Fastback, the GP pressure RMSE increases by a factor of 1.235 on Estateback, while its predicted standard deviation increases by 1.039, giving a growth ratio of 0.841. For concrete MC dropout, the corresponding factors are 1.183 and 1.029, giving 0.869. For the ensemble, they are 1.346 and 1.108, giving 0.823. A global calibration factor multiplies in-distribution and out-of-distribution uncertainty equally and therefore cannot change the ratios.

The methods also differ in their ability to identify unfamiliar geometries. In our evaluation, Notchback is treated as the smaller geometry shift because its overall rear-body morphology is closer to Fastback, whereas Estateback has a more distinct, bluff rear geometry. This ordering motivated training on Fastbacks and evaluating separately on Notchbacks and Estatebacks.

For Notchback, all AUROC values lie between 0.451 and 0.562, close to the chance value of 0.5. The mean predicted uncertainty therefore cannot reliably distinguish Notchbacks from the Fastbacks used for training. Estateback produces a clearer unfamiliarity signal: AUROC increases to 0.719 for GP total, 0.650 for concrete MC dropout, and 0.883 for the ensemble. The results are consistent with the intended ordering of the shifts: the smaller Notchback shift is difficult for every method to detect, while the larger Estateback shift is more readily identified.

Concrete MC dropout and ensemble disagreement estimate epistemic variation, so the GP epistemic component provides the more direct comparison. Its AUROC is 0.452 on Notchback and 0.684 on Estateback, compared with 0.514 and 0.719 when GP total uncertainty is used. This difference indicates that the learned residual term contributes to OOD separation as well as to interval magnitude.

After recalibration, the ensemble has the most stable interval magnitude under the body-style shifts and the strongest Estateback detection. 

\FloatBarrier
\subsection{Engineering Quantities of Interest}
\label{sec:drivaer-drag}

The surface pressure and wall-shear-stress fields are intermediate predictions; the engineering quantity considered here is the drag coefficient. We therefore examine how uncertainty in these fields propagates to uncertainty in drag. 

For a surface discretized into cells $i$, the drag coefficient is computed as

\begin{equation}
C_D
=
\kappa
\sum_i A_i
\left[
(\mathbf{n}_i\cdot\mathbf{d})p_i
-
\boldsymbol{\tau}_i\cdot\mathbf{d}
\right],
\label{eq:drag-integral}
\end{equation}

where $A_i$ and $\mathbf{n}_i$ are the area and normal of cell $i$, $\mathbf{d}$ is the drag direction, $p_i$ is pressure, $\boldsymbol{\tau}_i$ is wall shear stress, and $\kappa=2/(A_{\mathrm{ref}}\rho U_\infty^2)$ is the force-coefficient prefactor. Assuming independence across cells and physical channels, the corresponding variance is propagated as

\begin{equation}
\sigma_{C_D}^{2}
=
\kappa^2
\sum_i A_i^2
\left[
(\mathbf{n}_i\cdot\mathbf{d})^2\sigma_{p,i}^{2}
+
\sum_{j=1}^{3}d_j^2\sigma_{\tau_{ij}}^{2}
\right].
\label{eq:drag-variance-propagation}
\end{equation}

For the marginal calculation, the same propagation rule is applied to every method. For the GP-based method, $\sigma_{p,i}$ and $\sigma_{\tau_{ij}}$ are the predicted pointwise marginal standard deviations. For concrete MC dropout and the ensemble, they are estimated across stochastic passes or ensemble members.

The marginal rule omits spatial covariance, covariance between pressure and wall shear stress, and covariance between wall-shear-stress components. Across the 300 cases, sample-wise propagation changes the mean drag standard deviation by factors of approximately 1.1 for concrete MC dropout and 15--22 for the ensemble, depending on body style. These large scale changes do not by themselves establish improved identification of high-drag-error cases.

The sample-wise propagation retains the spatial and cross-channel dependencies represented by the sampled fields. An analogous sample-wise propagation for the GP-based method would require samples from its joint field posterior or an explicit covariance model, neither of which is exposed by the current pointwise-marginal interface.

The relationship between predicted drag uncertainty and absolute drag error is not consistent enough to rank the methods. Small changes in the integrated pressure and wall-shear-stress errors can reorder cases, while a shared bias in predicted drag creates an error component that no case-dependent uncertainty score can rank.

\subsection{Uncertainty-Guided Solver Review}

\begin{figure}[!htbp]
\centering
\includegraphics[width=\textwidth]{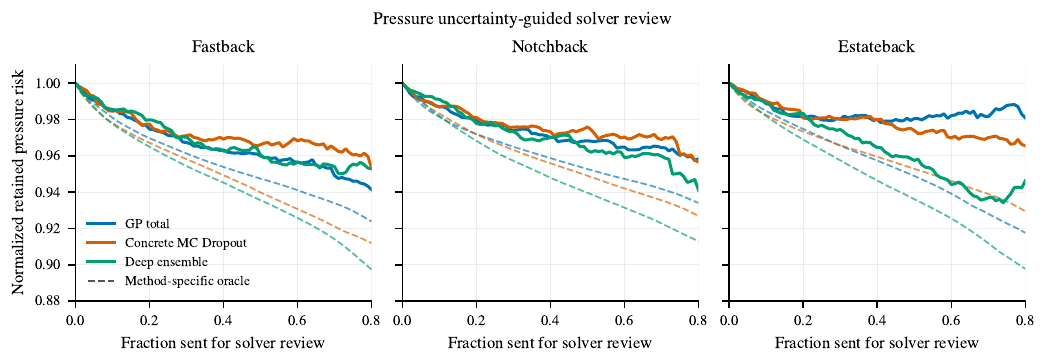}
\caption{Normalized retained pressure risk under uncertainty-guided solver review. Solid curves select cases using predicted uncertainty; dashed curves use the true case errors and show the oracle ordering.}
\label{fig:risk-coverage}
\end{figure}

Surrogate models can produce inaccurate predictions for some inputs. When only a subset of cases can be checked with the high-fidelity solver, a case-level uncertainty score can help prioritize which cases to review. Figure~\ref{fig:risk-coverage} ranks each body style by total GP standard deviation, concrete MC dropout disagreement, or ensemble disagreement. The score is the mean pressure standard deviation per case. At each review fraction, risk is the pressure RMSE over cases not sent back to the solver, normalized by the RMSE at zero review. For readability, the plot stops at a review fraction of 0.8, leaving at least 20 cases in each panel. The reported AUSE values are computed from the complete sparsification curves.

We summarize each ranking using pressure AUSE, the normalized area between the uncertainty-ranked and oracle risk curves over the complete review range. AUSE uses the full curves rather than their separation at any single review fraction; zero indicates oracle ordering. For GP total, concrete MC dropout, and the ensemble, respectively, pressure AUSE is 0.013, 0.026, and 0.030 on Fastback; 0.017, 0.021, and 0.022 on Notchback; and 0.037, 0.025, and 0.023 on Estateback. The 95\% intervals overlap in several comparisons, so the small differences are not decisive. Uncertainty provides some value for prioritizing cases with large pressure errors, but the improvement is modest.

\FloatBarrier
\subsection{Computational Trade-Offs}

The methods differ in where they incur their computational cost. Table~\ref{tab:drivaer-cost} summarizes the dominant requirements of the evaluated methods.

\begin{table}[!htbp]
\centering
\small
\begin{tabular}{lccc}
\toprule
\textbf{Method} & \textbf{Independently trained models} & \textbf{Stored checkpoints} & \textbf{Model evaluations per case} \\
\midrule
GP-based & 1 & 1 & 1 \\
Concrete MC dropout & 1 & 1 & 20 \\
Deep ensemble & 5 & 5 & 5 \\
\bottomrule
\end{tabular}
\caption{Evaluation counts for the DrivAerStar configurations.}
\label{tab:drivaer-cost}
\end{table}

The GP-based method requires one training run and one end-to-end model evaluation at inference. Its mean, epistemic variance, residual variance, and total predictive variance are produced in the same evaluation. The GP head evaluates inducing-point covariances in addition to the backbone. Training is more complex than deterministic regression because the backbone, variational distribution, inducing locations, kernel parameters, and residual-variance model are optimized jointly.

Concrete MC dropout also requires only one training run and stores one checkpoint. Its additional cost occurs primarily during inference, where $S$ stochastic passes are used for every case. The number of passes can be changed after training, allowing a trade-off between inference cost and the stability of the estimated sample variance.

The deep ensemble evaluated here contains $K$ independently initialized and trained members, stores $K$ checkpoints, and requires $K$ backbone evaluations per case. Both training and inference can be parallelized when sufficient hardware is available, reducing wall-clock latency but not total computation or storage. The complete field prediction from each member can be used for sample-wise propagation to nonlinear or spatially integrated engineering quantities. The DrivAerStar evaluation reports both sample-wise propagation and the common marginal-variance rule described in Section~\ref{sec:drivaer-drag}.

In the evaluated configurations, the GP shifts complexity to training, concrete MC dropout shifts additional cost to inference, and the ensemble multiplies training, storage, and inference work. A further distinction concerns the numerical stability of the predicted uncertainty field. For a fixed input and trained model, the GP-based method evaluates its marginal variance deterministically, without Monte Carlo estimation noise. Its regularity is governed by the learned feature map, GP kernel, and residual-variance model. Concrete MC dropout and ensemble standard deviations are instead empirical estimates from a finite number of field samples. Their estimates become more stable with additional stochastic evaluations or ensemble members. Under idealized independent Gaussian sampling, the relative standard error of an estimated standard deviation is approximately $1/\sqrt{2(S-1)}$ for $S$ samples. This gives approximately 16\% for the 20 concrete MC dropout passes and 35\% for the five ensemble members used here. These are theoretical pointwise sampling estimates rather than measurements of spatial roughness, and independently trained ensemble members need not behave as ideal Gaussian samples. Figure~\ref{fig:airfrans-dropout-convergence} provides a direct finite-sample illustration on AirFRANS.

\FloatBarrier
\section{Additional CAE Case Studies}
\label{sec:additional-case-studies}

We use AirFRANS and automotive crash as supporting case studies that change the shift axis, physical system, and engineering use of uncertainty. We only include GP-based method and concrete MC dropout in the subsequent studies. Their numerical values are therefore compared only within each dataset, and the cross-case analysis focuses on qualitative findings supported by compatible measurements.

\subsection{AirFRANS}
\label{sec:airfrans-results}

AirFRANS supplies the continuous angle-of-attack extrapolation study defined in Section~\ref{sec:experimental-design}. 

\paragraph{Predictive accuracy}

Table~\ref{tab:airfrans-accuracy} reports mean relative $L_2$ error across the four predicted channels together with pressure error, which is particularly relevant for aerodynamic quantities.

\begin{table}[H]
\centering
\small
\begin{tabular}{lcccc}
\toprule
\textbf{Method} &
\textbf{Mean ID} &
\textbf{Mean OOD} &
\textbf{Pressure ID} &
\textbf{Pressure OOD} \\
\midrule
Deterministic & 0.082 & 0.098 & 0.126 & 0.150 \\
GP-based & 0.085 & 0.150 & 0.074 & 0.144 \\
Concrete MC dropout & 0.181 & 0.191 & 0.173 & 0.205 \\
\bottomrule
\end{tabular}
\caption{AirFRANS relative $L_2$ error in distribution (ID) and under angle-of-attack extrapolation (OOD). Mean error averages the two velocity components, pressure, and transformed turbulent viscosity.}
\label{tab:airfrans-accuracy}
\end{table}

In distribution, the GP-based model has nearly the same mean error as the deterministic reference. Its pressure advantage narrows under extrapolation, from an error difference of 0.052 in distribution to 0.006 outside the fitting range. Meanwhile, its mean error increases from 0.085 to 0.150, whereas the deterministic model increases from 0.082 to 0.098. The accuracy behavior of this evaluated GP configuration therefore differs from its behavior on DrivAerStar.

Concrete MC dropout has higher error both in and outside the fitting range. Its OOD-to-ID error ratio is consequently close to one partly because its in-distribution error is already high. This is important when interpreting its uncertainty calibration: an interval can appear to transfer more successfully when the underlying prediction error changes little because the model underfits both splits.

\paragraph{Uncertainty under angle-of-attack extrapolation}

Table~\ref{tab:airfrans-uq} summarizes interval magnitude, error discrimination, and response to shift for the GP-based and concrete MC dropout models.

\begin{table}[H]
\centering
\small
\begin{tabular}{lccc}
\toprule
\textbf{Metric} & \textbf{GP total} & \textbf{GP epistemic} & \textbf{Concrete MC dropout} \\
\midrule
Raw ID $z$-RMS & 1.090 & -- & 1.183 \\
Calibrated OOD $z$-RMS & 4.657 & -- & 2.046 \\
OOD error--uncertainty Spearman $\rho$ & 0.682 & 0.638 & 0.625 \\
OOD AUROC & 0.849 & 0.914 & 0.773 \\
Growth ratio & 0.701 & 0.707 & 0.768 \\
\bottomrule
\end{tabular}
\caption{AirFRANS uncertainty metrics averaged over the four output channels. Calibration factors are fitted on the in-distribution validation split and applied unchanged to the angle-of-attack extrapolation cases.}
\label{tab:airfrans-uq}
\end{table}

\begin{figure}[!htbp]
\centering
\includegraphics[width=\textwidth]{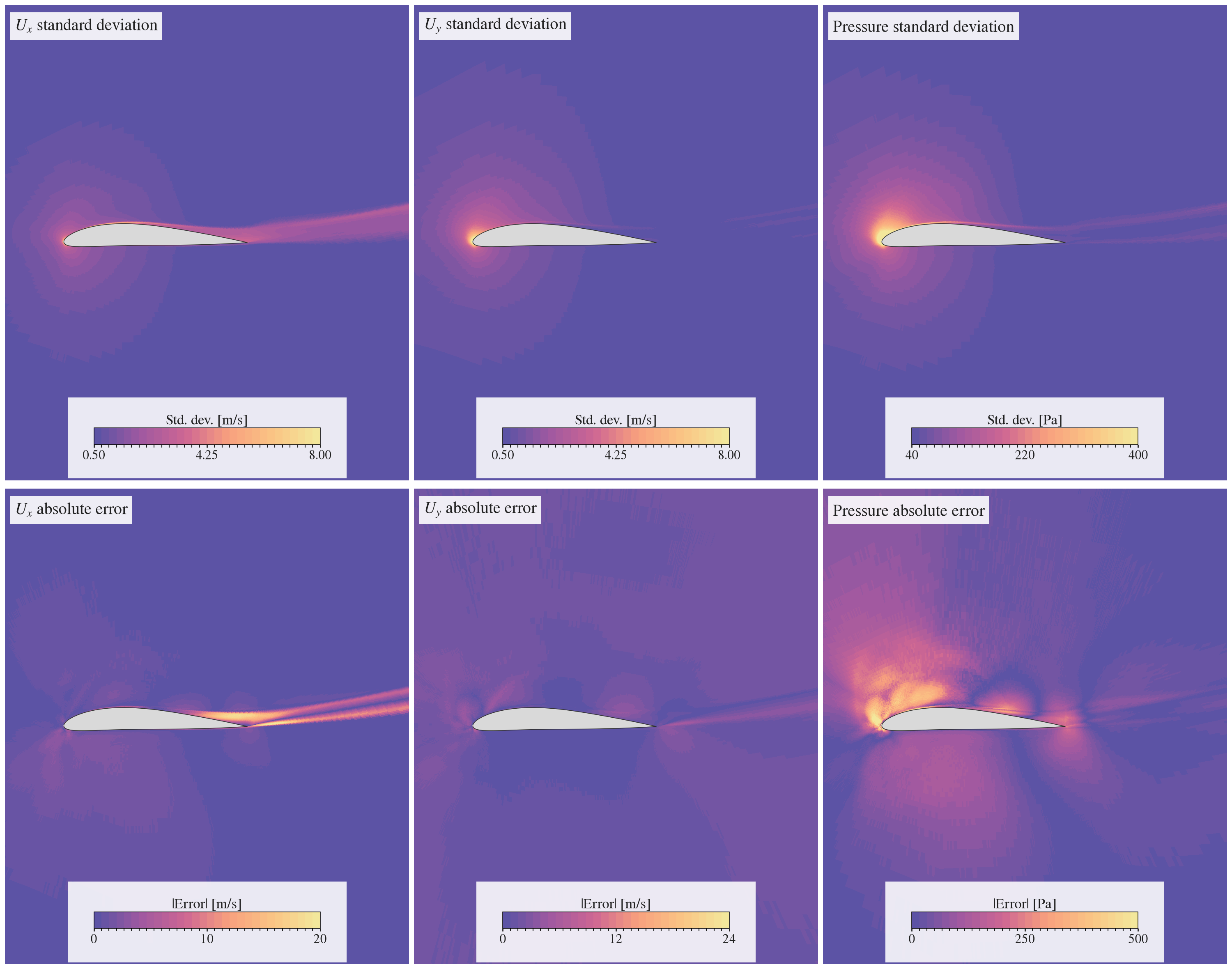}
\caption{GP total predictive standard deviation (top) and absolute prediction error (bottom) for $U_x$, $U_y$, and pressure on one representative AirFRANS extrapolation case at $\alpha=14.926^\circ$. The example visualizes spatial correspondence in a single flow field.}
\label{fig:airfrans-gp-fields}
\end{figure}

Figure~\ref{fig:airfrans-gp-fields} compares the GP uncertainty and absolute-error fields for a representative extrapolation case.

\begin{figure}[!htbp]
\centering
\includegraphics[width=\textwidth]{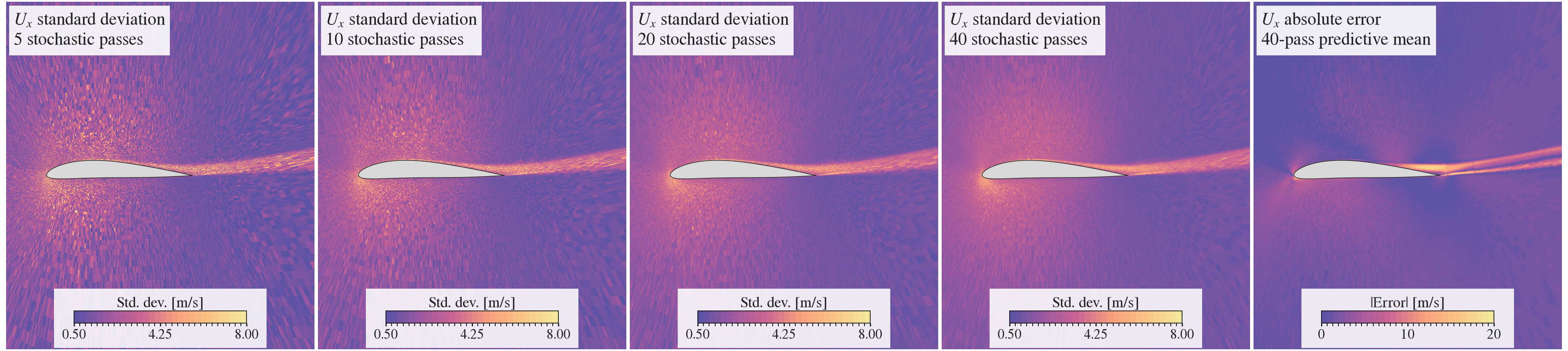}
\caption{Finite-sample behavior of concrete MC dropout on the same AirFRANS extrapolation case. The first four panels estimate the $U_x$ standard deviation from 5, 10, 20, and 40 stochastic passes; the final panel shows absolute error for the 40-sample predictive mean. Against this fixed error field, pointwise Spearman correlation increases from 0.570 to 0.623, 0.645, and 0.658. Rank agreement with the 40-sample standard-deviation field is 0.860, 0.942, 0.981, and 1.000, respectively.}
\label{fig:airfrans-dropout-convergence}
\end{figure}

Both raw uncertainty estimates are near nominal in distribution: the GP-based method has a $z$-RMS of 1.090 and concrete MC dropout 1.183. On DrivAerStar, by contrast, the uncalibrated sampling spread requires a substantially larger correction. The raw calibration gap is therefore dataset dependent.

After fitting scale factors in distribution, both methods become overconfident under extrapolation. The OOD $z$-RMS rises to 4.657 for the GP-based method and 2.046 for concrete MC dropout. Concrete MC dropout's smaller calibration degradation occurs while its mean prediction remains substantially less accurate.

For every method, predicted uncertainty increases by a smaller proportion than prediction error. The growth ratios are 0.701 for GP total, 0.707 for GP epistemic, and 0.768 for concrete MC dropout. This reproduces the DrivAerStar observation under a different and continuous shift axis: a scale fitted in distribution cannot correct an uncertainty estimate that responds too weakly as error increases outside the fitting range.

When the GP learned residual variance is excluded, GP epistemic uncertainty and concrete MC dropout disagreement have similar OOD error--uncertainty correlations, 0.638 and 0.625. GP epistemic uncertainty has higher OOD AUROC in every channel, averaging 0.914 compared with 0.773 for concrete MC dropout. GP total uncertainty has the highest field-error correlation, 0.682, but its OOD AUROC is lower at 0.849.

\FloatBarrier
\subsection{Automotive Crash}
\label{sec:crash-results}

The automotive crash study changes both the physics and the scope at which uncertainty is used. It uses the transient structural task and seven held-out runs defined in Section~\ref{sec:experimental-design}.

\paragraph{Mesh-level uncertainty}

We evaluate uncertainty across the complete displacement field. This field-level analysis asks whether the predicted uncertainty has a credible magnitude, both before and after calibration, and whether larger nodal errors receive higher uncertainty. The uncertainty metrics pool the three displacement components over the evaluated mesh nodes and 25 non-initial time steps for the seven held-out designs. Table~\ref{tab:crash-mesh-uq} summarizes the results on the seven held-out designs.

For each uncertainty definition, we fit one scaling factor per time step using the eight validation runs and a fixed 20,000-node mesh subsample. These scaling factors are applied unchanged to the seven held-out designs at both the mesh and probe levels. At each time step, the procedure adjusts uncertainty magnitude but does not change the ranking of runs or spatial locations.

\begin{table}[H]
\centering
\small
\begin{tabular}{lccccc}
\toprule
\textbf{Method} &
\textbf{Rel. $L_2$} &
\textbf{Raw $z$-RMS} &
\textbf{95\% cov.} &
\textbf{Spearman $\rho$} &
\textbf{Calibrated $z$-RMS} \\
\midrule
Deterministic & 0.195 & -- & -- & -- & -- \\
Concrete MC dropout & 0.223 & 3.919 & 0.487 & 0.347 & 1.216 \\
GP total & 0.371 & 1.564 & 0.783 & 0.207 & 1.229 \\
GP epistemic & 0.371 & 4.695 & 0.546 & $-0.103$ & 1.244 \\
\bottomrule
\end{tabular}
\caption{Mesh-level crash results on seven held-out designs. Spearman correlation is computed between nodal displacement error and predicted standard deviation.}
\label{tab:crash-mesh-uq}
\end{table}

GP total has the more credible raw interval magnitude, while concrete MC dropout has the more accurate mean and stronger spatial error ranking. GP epistemic uncertainty is too narrow before scaling and has a negative error--uncertainty correlation. After validation-based scaling, all three $z$-RMS values are similar. The comparison therefore changes with the intended use: raw interval scale favors GP total, whereas local error discrimination favors concrete MC dropout.

\begin{figure}[!htbp]
\centering
\includegraphics[width=\textwidth]{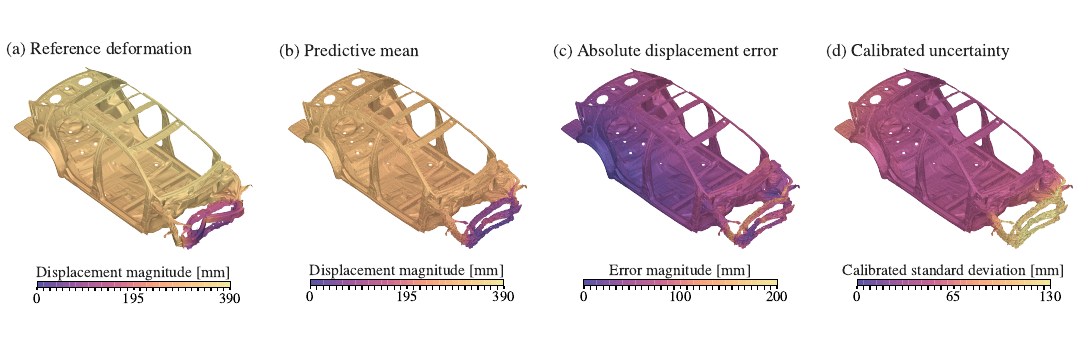}
\caption{Representative concrete MC dropout result for a held-out sample at 125\,ms. From left to right: reference deformation, predictive mean, absolute displacement error, and calibrated standard deviation.}
\label{fig:crash-cd-field}
\end{figure}

Figure~\ref{fig:crash-cd-field} visualizes the final state of a representative held-out crash simulation. Concrete MC dropout reproduces the overall deformation, while both the displacement error and calibrated uncertainty vary spatially and become most pronounced around the frontal impact structure.

\FloatBarrier
\paragraph{Probe-level uncertainty}

The engineering deliverable in the crash study is uncertainty at selected intrusion probes. We evaluate 185 probe nodes across the driver and passenger regions. For each method, we report both the uncertainty produced directly by the model and the uncertainty after post-hoc scaling.

Concrete MC dropout has approximately the same predictive accuracy as the deterministic reference, with RMSE values of 30.5 and 31.6\,mm, respectively. The GP-based method has a larger RMSE of 59.5\,mm, so interval metrics must be considered alongside this difference in mean accuracy.

The probe analysis uses the same validation-fitted, time-dependent scales described above. The probe nodes are not used during calibration.

Table~\ref{tab:crash-probe-uq} compares raw and calibrated probe uncertainty, pooled over the seven held-out runs and all 185 driver- and passenger-side probe nodes.

\begin{table}[H]
\centering
\small
\begin{tabular}{lccc}
\toprule
\textbf{Metric} & \textbf{Deterministic} & \textbf{Concrete MC dropout} & \textbf{GP total} \\
\midrule
RMSE (mm) & 31.6 & 30.5 & 59.5 \\
Raw sharpness (mm) & -- & 10.1 & 44.7 \\
Raw $z$-RMS & -- & 3.38 & 1.36 \\
Raw 95\% coverage & -- & 0.464 & 0.831 \\
Raw NLPD & -- & 8.89 & 5.52 \\
Calibrated sharpness (mm) & -- & 31.9 & 49.5 \\
Calibrated $z$-RMS & -- & 0.91 & 0.99 \\
Calibrated 95\% coverage & -- & 0.981 & 0.933 \\
Calibrated NLPD & -- & 4.70 & 5.13 \\
\bottomrule
\end{tabular}
\caption{Probe-level displacement uncertainty on seven held-out crash designs. Sharpness is the mean predicted standard deviation over the 185 probe nodes and 25 non-initial time steps. Coverage uses $\pm1.96\sigma$ intervals, with time-dependent scales fitted on the eight development runs.}
\label{tab:crash-probe-uq}
\end{table}

Before post-hoc scaling, GP total variance is closer to nominal: its raw $z$-RMS is 1.36 with 83.1\% coverage, compared with 3.38 and 46.4\% for concrete MC dropout. These values compare GP total uncertainty, which includes the learned residual variance, with concrete MC dropout disagreement, which represents epistemic uncertainty in this implementation. They therefore do not compare the two methods' epistemic uncertainties alone.

When calibration data are available, both methods approach the nominal interval magnitude. The GP-based method reaches a $z$-RMS of 0.99 with 93.3\% coverage, while concrete MC dropout reaches 0.91 with 98.1\% coverage. Concrete MC dropout retains the more accurate predictive mean and, after scaling, the lower NLPD: 4.70 compared with 5.13 for the calibrated GP. Before scaling, the ordering reverses, with NLPD values of 5.52 for the GP and 8.89 for concrete MC dropout.

\begin{figure}[!p]
\centering
\includegraphics[width=\textwidth]{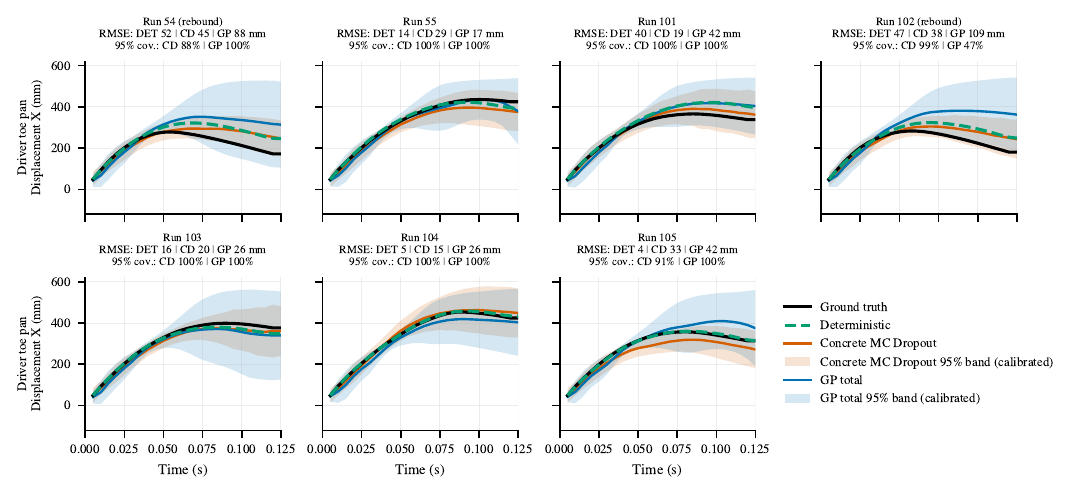}
\caption{Driver-side toe-pan displacement for the seven held-out crash designs, averaged over 104 probe nodes. Curves show ground truth and predictive means; shaded regions are calibrated 95\% Gaussian bands for both UQ configurations. Runs 54 and 102 have substantially larger GP errors without correspondingly wider intervals.}
\label{fig:crash-probe-runs}
\end{figure}

Within these seven held-out runs, aggregate calibration does not ensure that uncertainty responds to the most difficult designs. Two rebound-dominated runs have 3.54 times the GP prediction error of the other five runs, but their raw and calibrated uncertainty ratios are 0.99 and 1.00, respectively. For concrete MC dropout, the error ratio is 1.77, while its raw and calibrated uncertainty ratios are 0.89 and 0.83. In both methods, the predicted uncertainty remains approximately constant or decreases on the runs with larger errors. Figure~\ref{fig:crash-probe-runs} shows this behavior across all seven held-out designs.

\begin{figure}[!p]
\centering
\includegraphics[width=0.7\textwidth]{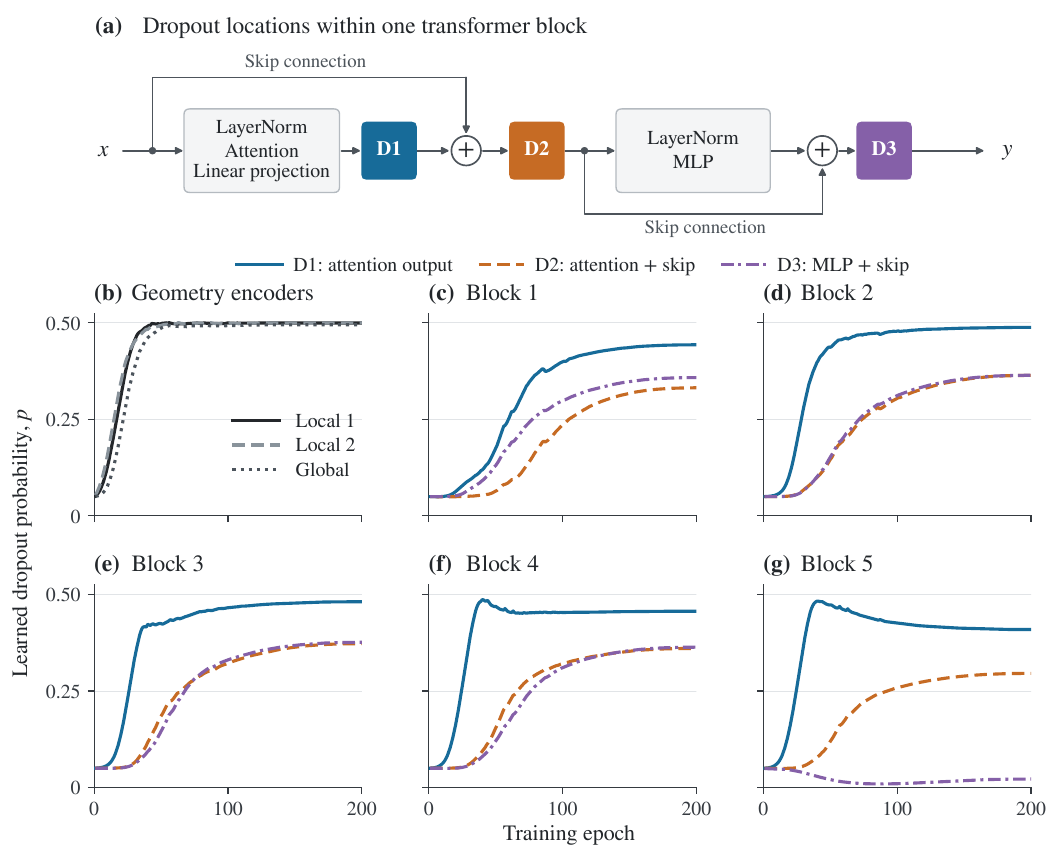}
\caption{Learned concrete dropout probabilities in the crash surrogate. (a) Dropout locations in one transformer block: D1 follows the attention-output linear projection, D2 follows the attention skip addition, and D3 follows the MLP skip addition. (b) Geometry encoders. (c--g) Transformer blocks, with matching colors and line styles for D1--D3. All rates were initialized to $p=0.05$.}
\label{fig:crash-dropout-rate-history}
\end{figure}

\paragraph{Learned dropout probabilities}
Figure~\ref{fig:crash-dropout-rate-history} tracks the 18 learned dropout probabilities in the crash model. The geometry-encoder rates approach 0.5 early, whereas the transformer-block rates evolve at different speeds; the final block's post-MLP residual rate remains small. The rates continue to change beyond the validation-selected checkpoint at epoch 30. Thus, checkpoint selection by predictive accuracy does not imply convergence of the dropout rates. These probabilities parameterize stochastic masks and do not measure layerwise predictive variance.

\FloatBarrier
\section{Cross-Case Synthesis and Practical Guidance}
\label{sec:cross-case-observations}

Table~\ref{tab:cross-case-summary} summarizes the main findings across the three datasets.

\begin{table}[!htbp]
\centering
\small
\begin{tabular}{p{0.28\textwidth}p{0.20\textwidth}p{0.20\textwidth}p{0.19\textwidth}}
\toprule
\textbf{Property} & \textbf{DrivAerStar} & \textbf{AirFRANS} & \textbf{Crash} \\
\midrule
Raw interval magnitude &
GP total variance closest to nominal; sampling disagreement too narrow &
Both methods near nominal &
GP total variance closest to nominal; Concrete MC dropout disagreement too narrow \\
\midrule
After ID scaling &
All nominal on unseen Fastbacks; transfer differs under shift &
Both overconfident under extrapolation &
Interval magnitudes closer to nominal on held-out runs
 \\
\midrule
Uncertainty growth under explicit shift &
All growth ratios below one &
All reported growth ratios below one &
-- \\
\midrule
Pointwise error discrimination &
All informative; GP total has highest Spearman correlation &
Both informative; GP has higher Spearman correlation &
Concrete MC dropout strongest; GP total positive; GP epistemic negative
 \\
\midrule
Case-level OOD detection &
Notchback near chance for all; Estateback ensemble strongest &
GP epistemic strongest; both GP scores exceed concrete MC dropout &
-- \\
\midrule
Engineering-quantity behavior &
Sample-wise propagation changes the estimated drag-uncertainty scale &

-- &
Probe uncertainty does not increase on the highest-error runs \\
\bottomrule
\end{tabular}
\caption{Cross-case summary. Comparisons are qualitative because the datasets use different targets, splits, aggregation rules, and available method sets.}
\label{tab:cross-case-summary}
\end{table}

The cross-case comparison suggests four practical lessons for evaluating UQ on other surrogate problems. First, the intended decision should determine the uncertainty quantity and evaluation metric. Figure~\ref{fig:drivaer-properties} shows that method performance changes across interval magnitude, local error discrimination, and OOD detection. The crash results provide another example: GP total variance has the more credible raw interval magnitude, whereas concrete MC dropout provides stronger mesh-level error discrimination (Table~\ref{tab:crash-mesh-uq}). Predictive intervals therefore require credible total uncertainty, error triage requires reliable ranking, and engineering quantities require uncertainty propagated through the quantity itself. Epistemic uncertainty is generally the relevant component for active learning. Success under one criterion should not be used as evidence for another.

Second, calibration corrects scale only under the conditions represented by the calibration data. On DrivAerStar, scale factors fitted on Fastbacks produce approximately nominal results on disjoint Fastback cases but transfer differently to Notchback and Estateback (Table~\ref{tab:drivaer-calibrated-magnitude}). On AirFRANS, scale factors fitted in distribution leave both methods overconfident under angle-of-attack extrapolation, and all reported growth ratios remain below one (Table~\ref{tab:airfrans-uq}). Because a positive scale factor cannot change uncertainty rankings or the relative growth of uncertainty between splits, a new application should reserve separate data for model selection, calibration, in-distribution testing, and testing under the shifts expected in deployment.

Third, field-level validation is not sufficient when the final decision uses an integrated force, extremum, probe response, or other derived quantity. In the DrivAerStar study, sample-wise propagation changes the mean drag standard deviation by a factor of approximately 1.1 for concrete MC dropout and by factors of 15--22 for the ensemble relative to marginal propagation (Section~\ref{sec:drivaer-drag}). This result shows that the treatment of spatial and cross-channel dependencies can materially affect the reported uncertainty scale. When a derived quantity matters, field samples or an explicitly joint predictive distribution should therefore be propagated through the same calculation used by the engineering workflow.

Finally, uncertainty should be evaluated alongside predictive accuracy and computational requirements. The crash probe results illustrate the first point: GP total variance has the more credible raw interval magnitude, but its RMSE is 59.5\,mm compared with 30.5\,mm for concrete MC dropout (Table~\ref{tab:crash-probe-uq}). The evaluated methods also require different numbers of trained models, stored checkpoints, and model evaluations per case (Table~\ref{tab:drivaer-cost}). An uncertainty estimate cannot compensate for an inaccurate predictive mean, and its training, storage, and inference requirements must be considered when selecting a method for deployment. Table~\ref{tab:uq-guidance} translates these findings into method-selection guidance.

\begin{table}[!htbp]
\centering
\scriptsize
\begin{tabular}{p{0.14\textwidth}p{0.22\textwidth}p{0.26\textwidth}p{0.26\textwidth}}
\toprule
\textbf{Method} & \textbf{Prediction interface} & \textbf{Well suited to} & \textbf{Checks required for a new problem} \\
\midrule
GP-based &
Posterior epistemic marginal variance and total marginal variance including learned model-discrepancy variance &
Single-pass local uncertainty, predictive intervals, and case-level scores; epistemic variance can support data acquisition &
Mean accuracy, interval calibration, response to relevant shifts, sensitivity to the learned features and GP configuration, and whether a joint distribution is needed for derived quantities \\
\midrule
Concrete MC dropout &
Coherent stochastic field samples; sample disagreement estimates epistemic uncertainty &
Spatially coherent uncertainty scenarios and propagation to derived quantities using one trained model &
Convergence with the number of stochastic passes, learned dropout probabilities, interval calibration, predictive accuracy, and response to relevant shifts \\
\midrule
Deep ensemble &
Field samples from independently trained models; member disagreement estimates epistemic uncertainty &
A strong sampling-based reference and deployment settings where multiple independently trained members and stored checkpoints are acceptable &
Member diversity, sensitivity to training initialization, convergence with ensemble size, interval calibration, predictive accuracy, and response to relevant shifts \\
\bottomrule
\end{tabular}
\caption{Method-selection guidance for extending the evaluated UQ approaches to a new surrogate problem. The appropriate choice depends on the downstream decision, required prediction interface, and available training and inference budget.}
\label{tab:uq-guidance}
\end{table}

No method is a default choice across these uses. For interval reporting, a new study should evaluate coverage, sharpness, and predictive accuracy on an independent test set after fixing any calibration procedure. For review prioritization or solver triage, it should evaluate error ranking and risk reduction directly; positive rescaling cannot improve these rankings. For OOD guardrails, it should test both detection and whether uncertainty grows at least as quickly as error across application-relevant shifts. For active learning, the epistemic component should be used and the acquisition strategy should ultimately be judged by the reduction in prediction error after new simulations are added.

Implementation choices should be validated as part of this process. Sampling-based estimates require a convergence study over the number of stochastic passes or ensemble members. GP-based estimates require sensitivity checks for the learned representation, inducing-point coverage, kernel constraints, and variance parameterization. UQ metrics should be tracked during training because their best checkpoints need not coincide with those selected for mean accuracy.

\FloatBarrier
\section{Limitations}
\label{sec:limitations}

The complete method matrix is available only on DrivAerStar; AirFRANS and crash compare the GP-based and concrete MC dropout configurations without a deep ensemble. The reported bootstrap intervals quantify sensitivity to the finite DrivAerStar test sets rather than to model training. Training-run sensitivity is evaluated separately for the GP-based and concrete MC dropout configurations on DrivAerStar, but not for the complete method matrix or the other datasets. 

The GP-based method reports learned residual variance in addition to posterior epistemic variance, while the sampling configurations report epistemic disagreement only. We therefore use total variance for predictive-interval evaluation and compare GP posterior variance with sampling disagreement for epistemic tasks.

The GP drag calculation uses pointwise marginals under an independence approximation because the current interface does not expose joint field samples. Concrete MC dropout and the ensemble are also evaluated with sample-wise propagation across all 300 cases. Post-hoc calibration is limited to positive scaling; conditional and conformal approaches are not evaluated.

Concrete MC dropout uses 20 stochastic passes on DrivAerStar and AirFRANS and 32 on crash; the DrivAerStar ensemble has five members. Their standard deviations contain finite-sample variation, illustrated in Figure~\ref{fig:airfrans-dropout-convergence}. The GP-based configuration avoids inference-time sampling noise but is sensitive to its feature representation, inducing-point coverage, kernel constraints, residual-variance bounds, and objective schedule.

\section{Conclusion}
\label{sec:conclusion}

We evaluated predictive uncertainty for neural CAE surrogates on three large, physically distinct problems: vehicle aerodynamics across body styles, airfoil-flow prediction under angle-of-attack extrapolation, and transient automotive crash dynamics. The datasets contain complex geometries, large irregular meshes, multiple spatial output channels, and engineering quantities of direct practical interest. This setting allowed us to examine UQ not only as a statistical output but as information that must survive the scale, geometry variation, and aggregation found in CAE workflows.

The experiments show that calibration, error discrimination, response to distribution shift, and engineering-quantity level usefulness are separate properties. A method that performs well on one need not perform well on another. In-distribution scaling corrects aggregate interval magnitude, but it cannot make uncertainty increase at the same rate as prediction error under distribution shift. Likewise, local correspondence between error and uncertainty does not guarantee useful case-level guardrails or reliable propagation to quantities of interest.

Both GP-based and sampling-based uncertainty remain useful within this picture. GP total variance provides a strong raw interval scale in the DrivAerStar and crash configurations. GP epistemic variance retains meaningful field-error discrimination and gives stronger epistemic OOD detection than concrete MC dropout on AirFRANS, while concrete MC dropout gives stronger mesh-level error ranking on crash. On DrivAerStar, sample-wise propagation of fields materially changes the estimated drag-uncertainty magnitude relative to marginal propagation. The methods' relative ordering therefore changes with the dataset and the engineering question, which is the central empirical result of this study.

The next step is to extend the framework with conformal prediction so that finite-sample coverage guarantees under exchangeability can be studied alongside the probabilistic metrics used here, including an evaluation of coverage under structured distribution shifts. We also plan to evaluate these methods on larger and more diverse industrial datasets and to move beyond passive uncertainty reporting. Important downstream directions include uncertainty-guided geometry modification, acquisition of new simulations through active learning, and design-space exploration in which uncertainty influences where and how a surrogate is queried. These workflows will test whether uncertainty can improve an engineering process, rather than only whether it scores well after prediction.

\section*{Acknowledgments}
We thank Sudeep Chavare for sharing the automotive crash dataset used in this study.

\bibliographystyle{unsrtnat}
\bibliography{refs}

\appendix
\renewcommand{\thesection}{Appendix \Alph{section}}
\section{Evaluated Configuration Details}
\label{app:configurations}

Table~\ref{tab:configuration-details} records settings that materially affect the interpretation or reproduction of the UQ results. The implementation source and experiment configuration remain the authoritative specification for layer-by-layer architecture and optimizer defaults.
\begin{table}[H]
\centering
\scriptsize
\begin{tabular}{p{0.14\textwidth}p{0.18\textwidth}p{0.59\textwidth}}
\toprule
\textbf{Dataset} & \textbf{Method} & \textbf{UQ-specific settings} \\
\midrule
DrivAerStar & GP & 1,024 inducing points per output; DKL widths 128/16; Matérn-$5/2$ ARD; radial $L_2$ feature normalization; residual head 64/64; residual-standard-deviation bounds 0.01--10; 200,000 geometry samples; 51,200-point GP inference chunks. \\
DrivAerStar & Concrete MC dropout & Learned concrete probabilities in the backbone; entropy coefficient $10^{-4}$; 20 stochastic passes; 200,000 geometry samples. \\
DrivAerStar & Ensemble & Five independent members with training seeds 1000--1004; member 0 is the deterministic reference; five member predictions at inference. \\
\midrule
AirFRANS & GP & 1,024 inducing points per output; DKL widths 128/16; residual head 64/64; 12,288 points per GP update; KL and GP-loss warmup through epoch 30. \\
AirFRANS & Concrete MC dropout & Entropy coefficient $10^{-4}$; matched GeoTransolver backbone and data resolution; 20 passes for manuscript evaluation. \\
\midrule
Crash & GP & One checkpoint for mesh and probe evaluation; 32,768 GP-objective nodes per time step; 512 inducing points; 128/16 DKL transform; 64/64 residual head. \\
Crash & Concrete MC dropout & Entropy coefficient $10^{-4}$; 32 inference passes; time-conditional GeoTransolver backbone with FLARE attention. \\
\bottomrule
\end{tabular}
\caption{UQ-specific settings of the evaluated configurations.}
\label{tab:configuration-details}
\end{table}

DrivAerStar and AirFRANS use AdamW with learning rate $10^{-3}$, weight decay $10^{-4}$, $(\beta_1,\beta_2)=(0.9,0.999)$, and a StepLR schedule with factor 0.5 every 100 epochs. Their GP configurations ramp both the KL coefficient and overall negative-ELBO weight over epochs 0--30, use a final KL multiplier of 0.5, a unit-weight mean-MSE anchor, and a 0.5-weight latent-distance penalty over 4,096 pairs. Crash uses Adam with learning rate decayed from $3\times10^{-4}$ to $10^{-6}$ and weight decay $10^{-4}$. Its GP schedules ramp the KL term over epochs 10--60 and negative-ELBO weight over epochs 5--40; the final KL and mean-MSE weights are one, and the latent-distance penalty has weight 0.1 over 4,096 pairs. All GP heads use gradient-norm clipping at 10.

\paragraph{Backbone configurations.}
The DrivAerStar and AirFRANS experiments use GeoTransolver backbones with 20 transformer blocks, hidden width 256, eight attention heads, and 128 physical-state slices. DrivAerStar uses six 32-dimensional local-geometry scales, producing a 448-dimensional per-point representation, whereas AirFRANS uses four scales and a 384-dimensional representation. DrivAerStar receives centered surface coordinates and normals together with air density and freestream velocity, and predicts pressure and three wall-shear-stress components. AirFRANS receives centered volume coordinates, a direction relative to the airfoil centroid, freestream velocity, and angle of attack, and predicts two velocity components, pressure, and turbulent viscosity. Both configurations sample 51,200 field points during training; their geometry branches use 200,000 DrivAerStar surface points and 768 AirFRANS airfoil points, respectively. The corresponding GeoTransolver backbones contain approximately 29.49 million and 21.66 million parameters.

The crash experiments use a time-conditional GeoTransolver with FLARE attention. This configuration has five transformer blocks, hidden width 256, eight heads, 128 global queries, and a 320-dimensional per-node representation. Its inputs are the undeformed coordinates, component thickness, and normalized query time, and its outputs are the three displacement components. The backbone operates on the full 384,862-node mesh. The selected GP configuration evaluates its variational objective on 32,768 sampled nodes per training step. The deterministic and concrete MC dropout configurations contain approximately 5.60 million parameters, while the complete GP-based configuration contains approximately 6.46 million. The backbone hyperparameters are shared across methods.

\paragraph{GP implementation.}
For each dataset, the backbone first produces a learned representation at every predicted field location. The evaluated GP configurations then apply a pointwise deep-kernel network with widths 128 and 16, followed by normalization that retains both feature direction and magnitude. This produces a 17-dimensional kernel input. Each physical output channel is modeled by an independent sparse variational GP with a Mat\'ern-$5/2$ ARD kernel and its own variational parameters and inducing locations. DrivAerStar and AirFRANS use 1,024 inducing points per output channel, while crash uses 512. The latent GP posterior supplies the predictive mean and epistemic variance. A separate two-layer, 64-unit network maps the same kernel features to an input-dependent residual-variance term, and total predictive variance is formed by adding this term to the GP epistemic variance. For crash, normalized time is included in the backbone input and is also appended to the backbone representation before the deep-kernel transformation.

\section{DrivAerStar Reproducibility Across Training Runs}
\label{app:seed-sensitivity}

To evaluate sensitivity to model training, we trained and evaluated the GP-based and concrete MC dropout configurations three times using independent random seeds. Each training run was calibrated separately on the same disjoint 100-case Fastback calibration set and then evaluated on the three 100-case test sets.

Table~\ref{tab:drivaer-multiseed} summarizes the variation in field-level accuracy and UQ metrics across these training runs.

\begin{table}[H]
\centering
\scriptsize
\begin{tabular}{lccccc}
\toprule
\textbf{Method} & \textbf{Pressure $L_2$} & \textbf{$z$-RMS F} & \textbf{$z$-RMS N/E} & \textbf{Epistemic $\rho$} & \textbf{Total AUSE} \\
\midrule
GP-based & $0.252\pm0.009$ & $1.005\pm0.005$ & $1.274\pm0.024$ & $0.620\pm0.008$ & $0.027\pm0.002$ \\
Concrete MC dropout & $0.221\pm0.002$ & $1.001\pm0.002$ & $1.121\pm0.003$ & $0.567\pm0.005$ & $0.038\pm0.002$ \\
\bottomrule
\end{tabular}
\caption{DrivAerStar sensitivity to model training. Values are mean $\pm$ sample standard deviation over three independent training runs. Pressure relative $L_2$ error, epistemic error--uncertainty Spearman correlation $\rho$, and total AUSE are first averaged over Fastback, Notchback, and Estateback. Calibrated $z$-RMS is shown separately for Fastback (F) and the average of the held-out Notchback and Estateback classes (N/E).}
\label{tab:drivaer-multiseed}
\end{table}

The field-level conclusions are consistent across the three training runs. Concrete MC dropout retains lower pressure error, while the GP-based method retains stronger field-error discrimination and lower total AUSE. Separate calibration brings both methods close to unit $z$-RMS on Fastback, but interval scale still degrades on the held-out body styles, more strongly for the GP configuration.

Table~\ref{tab:drivaer-drag-stability} evaluates whether drag and marginal drag-uncertainty rankings are similarly stable across training runs.

\begin{table}[H]
\centering
\scriptsize
\begin{tabular}{llccc}
\toprule
\textbf{Comparison} & \textbf{Method} & \textbf{Fastback} & \textbf{Notchback} & \textbf{Estateback} \\
\midrule
Predicted vs.\ true drag & GP-based & 0.946--0.967 & 0.932--0.946 & 0.861--0.881 \\
 & Concrete MC dropout & 0.955--0.961 & 0.921--0.935 & 0.888--0.917 \\
\midrule
Predicted drag across training runs & GP-based & 0.976--0.986 & 0.972--0.981 & 0.974--0.981 \\
 & Concrete MC dropout & 0.988--0.991 & 0.976--0.982 & 0.977--0.983 \\
\midrule
Marginal drag UQ across training runs & GP-based & 0.958--0.973 & 0.949--0.976 & 0.940--0.964 \\
 & Concrete MC dropout & 0.975--0.985 & 0.984--0.989 & 0.891--0.903 \\
\bottomrule
\end{tabular}
\caption{Stability of DrivAerStar drag and marginal drag-uncertainty rankings. Entries are ranges of Spearman correlation over three independent training runs for predicted-versus-true drag and over the three pairs of runs for between-run agreement.}
\label{tab:drivaer-drag-stability}
\end{table}

For both the GP-based and concrete MC dropout methods, the Spearman correlation between predicted and true drag exceeds 0.86 for every body style and training run. Predicted-drag rankings are also consistent between runs, with correlations above 0.97, and marginal drag-uncertainty rankings have correlations above 0.89. The association between drag uncertainty and absolute drag error is less consistent because small changes in integrated field residuals can reorder absolute drag errors.

\end{document}